\RequirePackage{fix-cm}
\documentclass[smallextended]{svjour3}       %
\smartqed  %
\usepackage{graphicx}
\usepackage{xcolor}
\usepackage{caption}
\usepackage{amssymb}
\usepackage{booktabs}
\usepackage{tabularx}
\usepackage{array,amsmath}

\usepackage{enumitem}%

\usepackage[hidelinks]{hyperref}

\usepackage[]{natbib}
\begin{document}

\title{Analyzing the Impact of Companies on AI Research Based on Publications
}

\author{Michael Färber         \and
        Lazaros Tampakis 
}

\institute{Michael Färber \at
              Institute AIFB, Karlsruhe Institute of Technology (KIT), Germany  \\
              \email{michael.faerber@kit.edu}           %
           \and
           Lazaros Tampakis \at
              Institute AIFB, Karlsruhe Institute of Technology (KIT), Germany \\
              \email{lazaros.tampakis@student.kit.edu}   
}

\date{Received: 9 November 2022 / Accepted: 23 October 2023}

\maketitle

\begin{abstract}
Artificial Intelligence (AI) is one of the most momentous technologies of our time. Thus, it is of major importance to know 
which 
stakeholders influence  AI research. 
Besides researchers at universities and colleges, researchers in companies 
have hardly been considered in this context. 
In this article, we consider how the influence of companies on AI research can be made measurable on the basis of scientific publishing activities. 
We compare academic- and company-authored AI publications published in the last decade and use scientometric data from multiple scholarly databases to look for differences across these groups and to disclose the top contributing organizations. 
While the vast majority of publications is still produced by academia, we find that the citation 
count an individual publication receives is significantly higher when it is (co-)\allowbreak{}authored by a company. Furthermore, 
using a variety of altmetric indicators, we notice that publications with company participation 
receive considerably more attention online. 
Finally, we 
place our analysis results in a broader context and present targeted recommendations to safeguard a harmonious balance between academia and industry in the realm of AI research. 
\end{abstract}

\section{Introduction}
\label{sec:introduction}

Artificial intelligence (AI) 
research is a rapidly growing field of study that is likely to revolutionize both industry and society in the upcoming decades \citep{Makridakis.2017}. Consequently, the leaders in AI research will play a significant role in shaping this revolution. It is therefore of great importance 
for 
academia, industry, government agencies, policy makers, and the general public 
to know which 
stakeholders 
have the most impact in this area. %

Traditionally, the academic field has focused primarily on basic research, education, and training, while the industrial field 
has engaged in 
applied research and development in commercially viable application areas. 
In the field of AI, however, these tendencies seem to have shifted in recent years \citep{Gil.07.08.2019,Littman.2021}.
Many technology companies 
have massively invested into AI related research and development 
\citep{Dernis.24.09.2019} and are now able to compete with the best academic research institutions---best illustrated by their increased presence in prestigious AI conferences such as the Conference on Neural Information Processing Systems (NeurIPS) and the International Conference on Machine Learning (ICML) \citep{Ahmed.22.10.2020,Hagendorff.2021}. This is particularly remarkable as companies seem to have lowered their publication efforts in other disciplines \citep{Lariviere.2018,Tijssen.2004}.
In addition, it has been observed that publication activities among companies became more concentrated \citep{Ahmed.22.10.2020,Krieger.2021}. 
For instance, large language models such as GPT-3 were developed and trained with an entire research team at OpenAI in the background. Training GPT-3 took around one month and costed tens of million US Dollar.\footnote{\url{https://www.nytimes.com/2020/11/24/science/artificial-intelligence-ai-gpt3.html}} It is easy to imagine that such resources are not readily available to researchers in academia. As a consequence, many academics and policymakers call for a democratization of AI.

Existing 
approaches to evaluating the impact of corporate AI research
fall short in several respects:
Some researchers only analyze the number of publications as a proxy for research participation \citep{Ahmed.22.10.2020,Zhang.09.03.2021} and hence disregard citations and the impact of research outputs. %
Other scholars rely on a single metric such as the average citation count \citep{Klinger.22.09.2020} or the median citation count \citep{Hagendorff.2021}. 
In fact, a study suggests that there is no significant quality difference between academic and corporate research outputs in AI \citep{Hartmann.2020}.
In addition, most scientists based their analysis on a limited set of pre-selected conference papers \citep{Ahmed.22.10.2020,Hagendorff.2021,Hartmann.2020} and are thus neither comprehensive nor representative. 
Given these shortcomings, it follows that it is still largely unknown how the research quality and impact of corporate and academic AI research compare in recent years.

This article aims to 
provide an in-depth assessment of the impact of AI research conducted by companies and to compare academia- with company-authored AI publications\footnote{In this article, we use the terms ``publication'' and ``(research) paper'' interchangeably.} using large scholarly datasets, such as the Microsoft Academic Graph,\footnote{\url{https://www.microsoft.com/en-us/research/project/microsoft-academic-graph/}} the associated Microsoft Academic Knowledge Graph,\footnote{\url{https://makg.org/}} the Altmetric.com database,\footnote{\url{https://www.altmetric.com/}} and the Global Research Identifier Database.\footnote{\url{https://www.grid.ac/}} 

Overall, we make the following contributions:
\begin{enumerate}
    \item We analyze the impact of AI research conducted by companies using large scholarly datasets and using a combination of citations and altmetrics.\footnote{Our source code and data is available online at \url{https://github.com/LazaTabax/AI-Impact-Scientometrics}.}
    \item We disclose key contributors and topics concerning company-involved AI research.
    \item We place our analysis results into a broader context and provide targeted recommendations for action for science, industry, and politics. %
\end{enumerate}

The upcoming sections are structured as follows:
In Section \ref{sec:relatedwork}, we outline related work. In Section \ref{sec:data}, we describe how we created our dataset for analyzing the impact of companies on AI research, before presenting our methodology for data analyis in Section \ref{sec:approach}. 
Section \ref{sec:results} presents the results of our analyses. In Section \ref{sec:discussion}, we discuss our findings. We conclude in Section \ref{sec:conclusion}.

\section{Related Work}
\label{sec:relatedwork}

In this section, we outline 
existing works on analyzing the impact of research quantitatively and qualitatively. 
We then describe how our analysis differs from existing works. 

\subsection{Analyzing Research Quantity} %
\label{subsec:RLquantity}

A report from the Joint Research Centre of the European Commission and the OECD 
identified the top corporate R\&D investors \citep{Dernis.24.09.2019}. Not only did they display which of them published the most papers related to AI in the period from 2014 to 2016, they also looked into which of them filed the most patents or registered the most trademarks. 
However, they did not 
put the contributions of these companies in relation to the contributions made by the academic field.

The annual Artificial Intelligence Index Report by Stanford University included such comparisons \citep{Zhang.09.03.2021}. Among other things, the authors analyzed peer-reviewed papers related to AI from Elsevier's Scopus database.
They note that in every major country and region, the highest proportion of peer-reviewed AI papers
comes from academic institutions. However, the second highest share differs across regions: In the United States, corporate-affiliated research represents 19.2\% of the total publications, whereas in China (15.6\%) and the European Union (17.2\%), government institutions rank second in terms of publications.
The authors also mention that corporate-academic collaboration has significantly grown in importance and popularity in the AI field \citep[p.~23]{Zhang.09.03.2021}. 
According to the authors, publications in AI research in general continue to grow at an increasing pace, and China overtook the United States for the share of AI journal citations for the first time. The US, on the other hand, are still ahead at AI conferences in terms of publications and citations \citep{Zhang.09.03.2021}.
The report provides a great overview over the field but does not delve further into the differences between corporate research and academic research.

\citet{Jurowetzki.2021} 
analyzed publication counts by company and non-company institutions. They detect an increased proportion of corporate publications in AI research in the 2010s. They conclude that companies hold an increasingly important position in fundamental AI R\&D. They collect evidence of the brain drain of researchers from academia to industry and find that researchers working in the field of deep learning and those with higher average impact are more likely to switch \citep[p. 25]{Jurowetzki.2021}.

The study settings of 
\citet{Ahmed.22.10.2020} allowed the authors to analyze such differences in more detail. They also collected data through the Scopus database, but they focused on papers published at major AI and non-AI conferences. To classify affiliations, they used a fuzzy string matching and regular expressions 
\citep[p. 14]{Ahmed.22.10.2020}. In addition, they differentiated between different company sizes and university rankings by combining their affiliation data with information from Fortune magazine and QS World University Rankings. 
Their findings show a clear upward trend in corporate representation at all major AI conferences, as well as increased cooperation between elite universities and companies since the advent of deep learning. This is remarkable since companies seem to have lowered their publication efforts in other disciplines \citep{Tijssen.2004,Lariviere.2018}. 
Large technology companies and elite universities, with which they often work, have clear advantages in modern AI research, according to their reasoning. They argue that the underlying reason for this trend is the computational advantages of large companies, which they refer to as the ``compute divide.'' As a result, middle to low tier universities are effectively crowded out in AI as they lack access to such computational resources. 
Although their analysis shows that there has been an increase in the participation of companies in AI research, this study only considers the publication quantity and does not evaluate the quality of the publications or more specifically their citation impact.

\subsection {Analyzing Research Quality} %
\label{subsec:RLquality}

Several researchers 
aimed for measuring differences in the research quality. 
\citet{Hagendorff.2021}, for instance, analyzed 10,000 machine learning papers published between 2015 and 2019 from three leading conferences, namely CVPR\footnote{Conference on Computer Vision and Pattern Recognition}, NeurIPS\footnote{Conference on Neural Information Processing Systems}, and ICML.\footnote{International Conference on Machine Learning} To distinguish between the different affiliation types, the authors searched for a predefined set of academic and industry words in each paper's full-text. 
Consistent with the previously discussed works \citep{Zhang.09.03.2021,Ahmed.22.10.2020}, the authors also state that the absolute number of papers and the proportion of academia-industry cooperation papers is rising across-the-board. More importantly, however, the authors aimed to compare 
the ``success'' or ``impact'' \citep[p. 8]{Hagendorff.2021} that papers had depending on their affiliation type. As an indicator for that, they used the median number of citations per publication. Their results show that there is a significant difference between academia and industry. Company papers from all considered years and conferences have higher median citations than purely academic ones. However, the effect seems to diminish over time, which is due to the fact that citations are typically slow to accumulate and thus generally lower for more recently published papers. Therefore, the authors admit that mere citation analyses may not be particularly credible for very recently published papers.
In addition to their simple citation analysis, the authors looked for mentions of trending topics in machine learning, such as ``adversarial,'' ``reinforcement,'' ``deep'' or ``convolution'' in the papers. This provided insights into how early and how often such topics were to be found in pure academic, pure industry, and mixed papers. As it turned out, academic papers seem to be lagging roughly two years behind in such mentions compared to company papers \citep[p. 8]{Hagendorff.2021}. A similar analysis of social impact terms 
did not yield a significant difference between the groups. 

\citet{Klinger.22.09.2020} considered the research trajectory and topic diversity of published AI research and inspected the influence of private sector organizations in AI research. They based their analysis on a sample of around 90,000 pre-prints that were published in AI-related categories on arXiv.org. Because arXiv does not hold curated records of affiliations or citations, they fuzzy matched the paper titles with the Microsoft Academic Graph and its authors with the Global Research Identifier Database. In contrast to other approaches, they only split their paper collection into two groups: papers with industry participation and papers without. 
They find that papers involving companies have higher median and average citations than those without company involvement, similar to Hagendorff's results \citep{Hagendorff.2021}. 
Remarkably, this observation holds true even when accounting for the publication year, the number of institutions involved, and the topical composition of the papers \citep[p. 31]{Klinger.22.09.2020}. 
Thus, the authors concluded that AI research undertaken by private companies is particularly influential.
In addition, they also noted that the top collaborators with private AI researchers are elite institutions in the US. %

\citet{Hartmann.2020} 
analyzed (meta-)data of 15,000 articles from five AI conferences, 
covering the period of 2004–2016. They then manually distinguished between corporate and non-corporate institution types and split the articles into groups accordingly. 
Their findings suggest that the scientific quality of corporate
publications is on average comparable to that of academic publications in AI, and that the increase in the number of corporate publications is not associated with a decrease in quality \citep[p. 9]{Hartmann.2020}. While the authors do not find significantly higher citation counts of industry papers as the two previously mentioned studies \citep{Hagendorff.2021,Klinger.22.09.2020}, 
the period they considered 
differs from the one in our work. It is possible that in more recent years the citation patterns have indeed changed as we will show in Sec.~\ref{sec:results}. 
Hartmann and Henkel further argue that the main reason companies engage in fundamental AI research is 
the central role that data plays in their business models. Large tech companies typically create and own major data assets which gives them a comparative advantage in conducting AI research over universities \citep[p. 372]{Hartmann.2020}.

\subsection{Current Shortcomings and Own Contribution}
\label{subsec:RLshortcomings}

Although the studies mentioned above have provided valuable insights into the research trajectory of AI, a comprehensive scientometric analysis of the rise of corporate science has not been performed so far. Many researchers base their claim of the rise of corporate AI research merely on increased publication counts by the industry or on increased academia-industry collaboration patterns \citep{Ahmed.22.10.2020,Hartmann.2020,Zhang.09.03.2021}. These quantity comparisons can be judged as helpful, but they mostly measure scientific participation and not impact.
Furthermore, the samples used in these analyses often consist of a relatively low number of publications taken from a few selected top conferences \citep{Hagendorff.2021,Hartmann.2020}, completely disregarding publications from journals and smaller conferences. Consequently, these datasets may not be sufficiently representative for the entire field of study.
Moreover, the performed impact measurements predominantly rely on average or median citation counts \citep{Hagendorff.2021,Klinger.22.09.2020}, 
while other citation metrics 
may better describe the citation distribution (see Sec.~\ref{sec:approach}). 
Furthermore, none of the existing works address the issue of time delay in citations. Particularly, recently published papers may not have had adequate time to accumulate citations, thereby hindering the observation of corresponding effects.

To address these issues and gain a deeper insight into how companies impact AI research, we work on the following improvements and extensions in this article: 
First, we incorporate considerably more venues and papers into our analysis while maintaining high data quality. 
Second, we better characterize the citation distribution by using other appropriate citation impact indicators. 
Third, we include altmetrics in our analysis because they 
capture 
impact 
in a 
timely manner and can provide a different perspective on impact.

\section{Dataset}
\label{sec:data}

In this section, we present the datasets used for analyzing the impact of companies on AI research. We outline our data sources in Sec.~\ref{subsec:data-sources} and the data generation process in Sec.~\ref{subsec:data-processing}.

\subsection{Data Sources}
\label{subsec:data-sources}

We used the following data sources for our analysis: 
\begin{itemize} %
    \item the Microsoft Academic Graph\footnote{\url{https://www.microsoft.com/en-us/research/project/microsoft-academic-graph/}} (MAG),
    \item the associated Microsoft Academic Knowledge Graph\footnote{\url{https://makg.org/}} (MAKG),
    \item the Altmetric.com database\footnote{\url{https://www.altmetric.com/}} (ALTM), and
    \item the Global Research Identifier Database\footnote{\url{https://www.grid.ac/}} (GRID).
\end{itemize}

In the following, we describe these data sources in detail.

\subsubsection{Microsoft Academic Graph}
\label{subsec:mag}

The Microsoft Academic Graph (MAG) \citep{Wang.2020} is a large, heterogeneous graph containing metadata about millions of scholarly entities, such as publications, authors, institutions, and fields of study.  We chose the MAG\footnote{We used the latest MAG data dump version that was available at the start of our data analysis (July 2021).}\textsuperscript{,}\footnote{OpenAlex (\url{https://openalex.org/}) is a current attempt to maintain and improve the data previously available at the MAG. However, at the time of our data analysis, OpenAlex was not yet available to the extent that it is today. According to our research, OpenAlex has no significant improvements in AI paper coverage.} over other scholarly databases for the following reasons: 

First, the MAG covers more publications than comparable databases like Scopus, Web of Science, Dimensions, and CrossRef, as a recent study confirmed~\citep{Visser.2021}. The used MAG version 
contains metadata of more than 260 million publications across all scientific domains and nearly 27 million computer science papers specifically. Unlike databases like Scopus, which mainly focus on peer-reviewed journal articles, the MAG also contains many conference proceedings, book chapters, patents etc. from the last few decades. To study the AI discipline, in which researchers often prefer to publish in conferences or online-archives rather than journals, a broad coverage of documents in the data is valuable \citep{Visser.2021}.

Second, the MAG provides bibliometric data suitable for citation analysis. Its coverage of citations is better than that of Scopus and Web of Science and comparable to that of Google Scholar \citep{Harzing.2017}.
The main advantage of using the MAG over Google Scholar for a full-fledged bibliometric analysis is that the MAG also provides rich metadata about authors and affiliations \citep{Hug.2017}.

\subsubsection{Global Research Identifier Database}
\label{subsec:grid}

The Global Research Identifier Database (GRID) is an openly available, global database of research-related organizations.\footnote{Similar to Microsoft Academic, the GRID is not being continued from 2022 on. It has also received an open-access successor database which built on its data named Research Organization Registry (ROR).} It provides unique and persistent identifiers, along with metadata, for more than 100,000 curated records of organizations. %

Our MAG data features GRID identifiers for most affiliations. This allows us to connect affiliations with their respective GRID entries.
Linking the MAG affiliations to GRID-IDs is important for our analysis, because it enables us to retrieve the organization type of given affiliations---knowing if an institution belongs to academia or industry. 
GRID distinguishes between \textit{education}, \textit{healthcare}, \textit{company}, \textit{archive}, \textit{nonprofit}, \textit{government}, \textit{facility}, and \textit{other} as types of organizations.\footnote{\url{https://www.grid.ac/pages/policies}}
For our analysis, we focus on the entries labeled as either \textit{company} or \textit{education}:
\begin{itemize}
    \item Research producing institutions are categorized under \textit{education} if they have the ability to grant degrees. This typically includes faculties, departments, and schools. They represent academia. 
    \item \textit{Company} organizations, on the other side, are defined as private business entities with the aim of gaining profit. They represent the industry. 
\end{itemize}

Since links between the MAG and GRID already exist, 
manual mapping or string matching, as required in other related work, is unnecessary. 

\subsubsection{Altmetric Database}
\label{subsec:altmetrics}

The Altmetric database
is one of the 
largest 
altmetric data providers currently available \citep{RobinsonGarcia.2014}. It contains high quality data related to mentions of scientific publications on the web, including social media. %
Operating since 2012, altmetric.com has been monitoring a range of non-traditional sources.
Today, the Altmetric database contains 191 million mentions of over 35 million research outputs (including journal articles, datasets, images, white papers, reports and more).\footnote{\url{https://www.altmetric.com/about-our-data/how-it-works-2/}} 
The sources include public policy documents, blogs, mainstream media articles, faculty opinions, and social media posts. Altmetric.com also collects the number of online readers on platforms such as Mendeley. However, these values are not used for the attention score. 

The Altmetric attention score is a sophisticated score based on all online mentions a research output has received from these sources. It is automatically calculated and weights mentions depending on three main factors: volume, sources, and authors.\footnote{\url{https://www.altmetric.com/about-our-data/the-donut-and-score/}} 
First, the score increases with the number of people mentioning a given publication. 
Notably, only one mention per person per source is counted to prevent individuals with multiple postings about a paper influencing the score. 
Second, mentions from different sources are weighted differently. To illustrate, a newspaper article contributes more than a blog post which contributes more than a tweet. 
Third, the author of an online mention also plays a role in how it contributes to the score. Altmetric.com looks on the frequency someone mentions scholarly articles, whether there is any bias towards a particular journal or publisher, and who the audience is. For instance, a physician sharing a link to a study with other physicians counts significantly more than a journal account pushing the same link automatically. 
\citet{RobinsonGarcia.2014} conclude that Altmetric.com is a transparent, rich and accurate tool for altmetric data. For our analysis, we rely on the Altmetric API provided for research purposes.

\subsubsection{Microsoft Academic Knowledge Graph}
\label{subsec:makg}

The Microsoft Academic Knowledge Graph (MAKG) is a large dataset with information about scientific publications and related entities \citep{Farber.2019}. It was built on the basis of the MAG data and contains additional and revised metadata about scholarly entities \citep{Faerber.2022}. 
MAKG's revised version features tags for publications, offering more detailed information about the respective areas of research. Extracted from the publications' abstracts and filtered using the TextRank algorithm, these tags 
extend the field of study labels given in the MAG. They enable us to glean additional insights into the research focus of given AI papers.

\subsection{Dataset Creation}
\label{sec:method}
\label{subsec:data-processing} 

\begin{figure}[tb]
  \centering
  \includegraphics[width=0.95\textwidth]{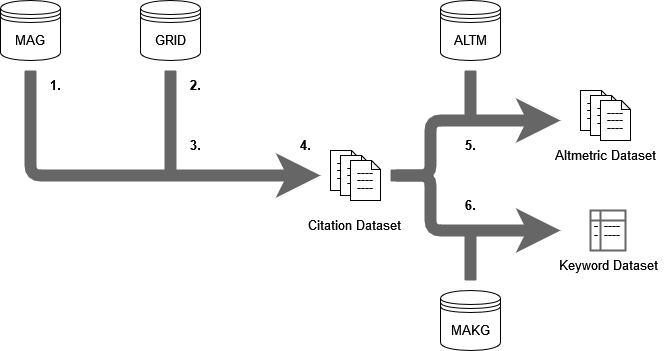}
  \caption{Schematic overview of the dataset creation pipeline} %
  \label{fig:22}
\end{figure}

A schematic overview of our data generation approach is given in Fig.~\ref{fig:22}. 
Overall, 
we perform the following steps: 
(1)~identifying relevant papers,
(2)~merging paper and affiliation metadata,
(3)~grouping papers into three disjoint groups, %
(4)~applying venue and time filters,
(5)~enriching the resulting sample with altmetric information,
(6)~and retrieving keywords from it. 
In the following, we describe these steps in more detail.

\begin{enumerate}
\item \textit{Identifying relevant papers:}  
The primary objective of the initial step is to gather metadata for relevant publications. Since our focus is on analyzing papers related to AI, we need to identify the papers within the MAG database that pertain to AI research. However, there are two challenges in determining which publications fall under the broad scope of AI.

Firstly, AI research intersects with various other scientific disciplines, making it difficult to establish clear boundaries. For example, disciplines like neuroscience or statistics share overlapping areas with AI, and the distinctions may not be well-defined. Secondly, it is unclear whether papers that utilize AI methods without directly contributing to AI knowledge should be included in our analysis. Some publications in fields such as finance or contemporary art may employ existing machine learning techniques without making significant contributions to the AI domain itself.

Every paper in the MAG database is already associated with one or more fields of study, thanks to Microsoft's assignment procedure outlined by \citet{Sinha.2015}. The assignment involved using ``seed'' papers as a starting point, which were either known to belong to a specific field or had relevant keywords in their titles. Additional papers were then identified as candidates based on their similarity to the top N related papers within a given field. The confidence score of related papers within a specific field increased based on their interconnectedness with other papers in that field.

The described assignment procedure and the possibility of assigning multiple fields of study to a paper help resolve the two challenges mentioned earlier. Assigning multiple fields of study to a paper enables us to include papers that belong to AI as well as other relevant domains such as mathematics. Additionally, the counting of connections a paper has to other papers within a particular field addresses the second challenge. This means that a paper that primarily applies AI methods to unrelated topics should either be excluded from the candidate list or assigned a very low confidence score.

Utilizing the MAG's integrated field of study categorization, we can specifically employ the field of study 'Artificial Intelligence' within the field of study 'Computer Science'. It is the sixth largest second-level field, comprising over 4.4 million papers. However, for increased data quality, we chose to exclude papers with a confidence score\footnote{This score ranges between zero and one and indicates the confidence of the field of study assignment, see \url{https://docs.microsoft.com/en-us/academic-services/graph/reference-data-schema\#paper-fields-of-study}.} of exactly zero, as this helps avoid the inclusion of potentially misassigned papers in the AI field.\footnote{The confidence values between 0 and 1 are not evenly distributed. Among the 520,000 articles with a confidence value greater than zero, 99.998\% have a value between 0.2 and 0.5. Therefore, we consider a threshold of 0.5 inappropriate for our scenario. Even if we were to set a threshold of 0.2, only 10 additional papers would be included in the sample.}

\item \textit{Merging paper and affiliation metadata:} 
In a second step, we combine the metadata from the identified papers with the metadata about the institutions of the papers' authors. 
As described in Sec.~\ref{subsec:mag}, the MAG 
assigns affiliations to authors and authors to papers. Hence, we can link papers via authors to affiliations (see Fig.~\ref{fig:2}). 
Notably, more than half of the (co-)\allowbreak{}authors (and thus about a quarter of the papers) had no information on their affiliation specified in the MAG; also, about 20\% of authors who had an affiliation entry cannot be mapped to a corresponding GRID entry. 
Ultimately, we obtain 7,700 unique affiliations associated with AI-research, linked to 315,000 papers, which is about 60\% of the papers from Step~1.

\begin{figure}
  \centering
  \includegraphics[scale=0.42]{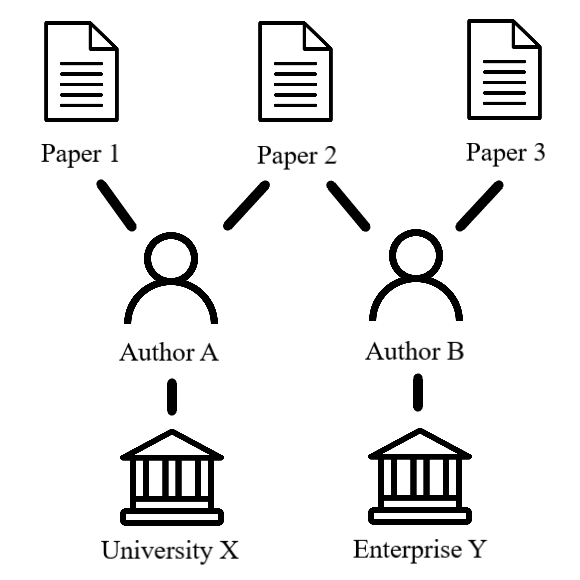} %
  \caption{Exemplary paper-author-affiliation combinations}
  \label{fig:2}
\end{figure}

\item \textit{Grouping papers into three disjoint groups:} 
Third, we match the metadata of publications and affiliations with the information about the organization types found in GRID. This allows us to split institutions (and thus papers) into a \textit{company} group and an \textit{education} group. About 5,500 of the 7,700 research organizations were identified as education types, 530 as company types. %
Note that there are plenty of papers co-written by authors from companies and from academia that can be found in both groups. This unclear distinction might skew results in either direction. Hence, we decided to put all papers that could be found in both groups into a separate third group. Similar to \citet{Hagendorff.2021}, we call this mixed group the \textit{cooperation} group, which leaves us with three disjoint groups of papers.
In the simplified examples of Fig.~\ref{fig:2}, paper 1 would be considered an \textit{education} paper, paper 2 a \textit{cooperation} paper, and paper 3 a \textit{company} paper.\footnote{If an author is employed at both a university and a company, we considered both affiliations and categorized all of these papers as collaborative papers. Upon reviewing our dataset, we discovered that only 18 papers had an author with both a corporate and an educational affiliation. However, since these papers included other (co-)authors with one of the respective affiliations, they would have been classified as collaborative papers regardless.} 

\item \textit{Applying venue and time filters:} 
Before performing our citation analysis, we filter the remaining 230,000 papers in two ways.
First, we exclude all papers not published in journals or conferences. This is necessary in order to ensure comparability across the different proportions of document types across groups.
Academic institutions almost exclusively publish journal articles and conference proceedings (over 90\% of papers). Companies' publications on the other hand include a significant amount of patents among other things (over a third of their documents). Since journal and conference papers have higher citation rates than other types of documents, this filter should in theory prevent this disparity potentially benefiting academia.

Second, we additionally exclude papers which were published before 2012. This has several reasons. The simplest one is that our main focus lies on the last decade of AI research. %
Additionally, recent developments such as the deep learning hype have their origins in the ImageNet Challenge from 2012 
\citep{Alom.03.03.2018,Ahmed.22.10.2020}. Moreover, a practical reason for choosing only papers from 2012 on is that Altmetric has only been established and has collected data from this year onwards. 

We call the resulting filtered dataset
the \textit{citation dataset} (see Fig.~\ref{fig:22}).
It 
consists of about 72,000 pure education papers, 3,800 cooperation papers, and 1,500 pure company papers.

\item \textit{Enriching the resulting sample with altmetric information:} 
Furthermore, based on the papers from the previous dataset, we create an
\textit{altmetric dataset} with altmetric indicators included (see Fig.~\ref{fig:22}). 
For integrating the altmetric data into this dataset, we query the Altmetric API 
and match about 17\% of the selected papers using their DOI. The relatively low matching proportion is to be expected and is in line with previous attempts \citep{RobinsonGarcia.2014} due to the fact that Altmetric does not create entries for unnoticed papers. 
This approach enables us to study the similarities and differences between citation-based indicators to altmetric ones for the considered AI papers.

\item \textit{Retrieving keywords:} 
At last, we access paper tags 
from the MAKG. Since the MAKG features up to five tags per paper, we are able to compile a list of over 300,000 tags attributable to papers from our \textit{citation dataset}. As these tags can be interpreted as papers' keywords, we name the resulting dataset \textit{keyword dataset} (see Fig.~\ref{fig:22}).
We also prepare two other paper samples through Step 1 to 4 to compare their citation distributions. %
The first one consists of AI papers published in five major AI conferences as in 
\citet{Hartmann.2020}, to reproduce and compare with the results of our main citation dataset. In the following we refer to it as the \textit{conference sample}.
Secondly, we are interested in non-AI papers to examine whether observed patterns are unique to the field of AI. For this, we perform the same data processing steps to papers from the unrelated field of electrical engineering. We refer to it as the \textit{non-AI sample}. 
\end{enumerate}

\section{Data Analysis}
\label{sec:approach}

In our data analysis, we use several measurements for quantifying the impact of AI research that we want to explain in greater detail in this section. First of all, we show how we approach the citation analysis in Sec.~4.1. In Sec.~4.2, we describe our use of altmetrics to measure a different dimension of impact. Finally, Sec.~4.3 outlines our method to analyze common AI research topics of academia and the industry. We conduct the analysis outlined in this section using Python, utilizing 
in particular the libraries  
pandas, scipy, and pyaltmetrics.

\subsection{Citation Analysis}

We start the citation analysis with comparing basic citation-based metrics like average citations and average citations per year that are calculated for each of the three groups \textit{education}, \textit{company}, and \textit{cooperation} independently. For measuring the differences between groups more appropriately, we apply percentile ranks and derive a weighted score from them. To validate the statistical significance of our results, we also apply significance tests. At last, we highlight top research organizations and their relative contributions in terms of publications and citations. 

In the following we explain the statistical methods used in the analysis: In Sec.~4.1.1 we briefly mention our outlier detection method, in Sec.~4.1.2 we discuss the percentile rank approach, and in Sec.~4.1.3 we explain the significance tests in use.

\subsubsection{Outlier Detection}
Citation distributions are typically highly skewed, meaning that the vast majority of papers receive little to no citations while a small fraction of successful papers gets cited very frequently.
To analyze whether such very successful papers distort our citation metrics, we for once exclude outliers and look into the metrics of the remaining papers. 
Following a standard method for detecting statistical outliers \citep{Outliers2019}, we define upper outliers as being above $Q3 + 1.5*IQR$, with $Q3$ being the 75th quantile and $IQR$ the interquartile range -- the difference between the 25th and 75th percentile in the sorted list of a given sample. In our \textit{citation dataset}, this applies to about 17.5\% of papers which received more citations than the group-specific outlier threshold.

\subsubsection{Percentile Ranks}
\label{sec:percentile-rank}
As citation distributions are typically highly skewed, arithmetic averages alone are not well suited for citation analyses.
Scientometricians argue, therefore, that \textit{percentiles} are an important alternative to such average-based indicators for obtaining a normalized citation impact of publications \citep{Bornmann.2013}.
With \textit{percentile ranks}, the citation score of each paper is rated in terms of its percentile in the total citation distribution—the 0th percentile being the lowest and the 99th percentile being the highest in citations. One can obtain percentiles from a set of numerical data by arranging it in ascending order and then splitting it into 100 groups of the same size.

There are two variants of the percentile rank approach: 
one with 100 rank classes 
(PR(100)), and one with 6 rank classes (PR(6)). 
The PR(6) method with 6 classes \citep{Leydesdorff.2011} rewards highly cited papers stronger than the PR(100), which is desirable for highly skewed distributions.
The PR(6) score 
has been widely used, such as by the National Science Foundation \citep{Leydesdorff.2011}. 
It also has the highest correlation with conventional indicators. For these reasons, we use 
it for the analysis.

To apply the PR(6) method, we split our paper collection into six classes: %
\begin{itemize}
    \item Rank 1: Bottom–50\%: papers with a percentile less than the 50th percentile;
    \item Rank 2: 50th–75th: papers within the [50th; 75th) percentile interval;
    \item Rank 3: 75th–90th: papers within the [75th; 90th) percentile interval; 
    \item Rank 4: 90th–95th: papers within the [90th; 95th) percentile interval; 
    \item Rank 5: 95th–99th: papers within the [95th; 99th) percentile interval; 
    \item Rank 6: Top-1\%: papers with a percentile equal to or greater than the 99th percentile.
\end{itemize}
Each paper that we want to analyze is assigned to one of these classes. The thresholds that define the class intervals are calculated from the aggregation of all papers.
Afterwards, each set of papers receives a mean percentile rank score by weighting the relative frequencies (i.e., probabilities) $p(x)$ in each set with their class rank $x$, as follows:
\begin{equation}
    R(6) = \sum_{x=1}^{6} x * p(x)
    \label{formula1}
\end{equation}
This means that papers that are, for instance, in the 80th percentile weigh three times as much as papers in the 40th percentile. 
The lowest possible score a set can receive is 1, for the case when all papers are in the bottom class. 
The highest score is 6 in the event that every paper gets placed in the top-1\% class. 

With this score we can evaluate whether the citation count of a given subset of papers is above or below a certain level. %
The expected value for the case of random attribution equals:  %

\begin{equation}
    1\cdot0.50 + 2\cdot0.25 + 3\cdot0.15 + 4\cdot0.05 + 5\cdot0.04 + 6\cdot0.01 = 1.91
    \label{formula2}
\end{equation}

\subsubsection{Statistical Significance Tests} 
To test the statistical significance of found differences, we use non-parametric tests, since they do not require us to make assumptions about the distributions at hand (e.g., assumption of a normal distribution or homogeneous variance in the subsamples).  Non-parametric tests, as suggested in 
\citet{Leydesdorff.2011}, work well with citation distributions, which are typically highly skewed.
First, we test whether differences among the subsets under study are significant using the Kruskal–Wallis rank variance test \citep{WilliamH.Kruskal.1952}. If the null hypothesis is not rejected (i.e., no significant differences among the sets are found), then the analysis ends because it is not relevant to test further.
In the other case, one can test for pairwise differences using the Mann–Whitney's U test \citep{MannWhitney1947} on each two subsets. We define a difference to be statistically significant when the p-value of the Kruskal-Wallis test is lower than the $\alpha = 0.05$ level, or lower than the Bonferroni adjusted $\alpha = 0.0167$ level when performing the three pairwise Mann-Whitney U tests on the subsets.\footnote{We also 
evaluate if the absence of keywords (i.e., not fields of study) would pose an issue, considering both the number of affected articles and any bias in company involvement between articles with and without keywords. Our examination reveals that articles lacking keywords accounted for a mere 0.04\% of the total articles we analyze. Additionally, these articles demonstrate a comparable level of company involvement to those with keywords, with an average company involvement of 7.4\% compared to 9.4\%. Based on these findings, we assert that the data used in our study is representative and reliable. 
To address concerns regarding potential variations in keyword semantics and their impact on our results, we can emphasize three key points. Firstly, our study focuses exclusively on AI as a restricted domain, minimizing the need for disambiguation of words. Secondly, we have identified only eight distinct AI subfields, such as \textit{machine learning} and \textit{AI ethics}, 
having rather non-overlapping topics. 
Thirdly, we employ a diverse range of keywords, including synonyms, for each AI subdomain.}  

\subsection{Altmetric Analysis} 
We follow a similar procedure with the attention scores of the matched paper sample from the Altmetric database. 
First, we analyze the basic statistics of the attention scores.
Second, we apply the PR6-method and test for statistical differences across the groups.
Third, we compare citation counts, attention scores and other altmetrics of papers to investigate the relationship between them.
Besides correlation, we are interested in the question of whether either group has some inherent bias towards more attention from fellow scholars (expressed by citations) or more attention from the public on the internet (expressed by altmetric attention score). %
To analyze this, we use the framework proposed by 
\citet{Williams.2022} that categorizes papers into three 
cases given their relative strength in the mentioned metrics: 
\begin{itemize}
    \item \textit{Exceptionals}: 
    highest scores in both citation count and altmetric attention score;
    \item  \textit{Scholars}: top in citations and lowest in altmetrics; and
    \item \textit{Influencers}: top in altmetrics and lowest in citations. 
\end{itemize}
Williams defined papers as top performing on a given metric if they land in the top 20\%; conversely, low performing papers on a given metric are papers in the lowest 20\%. 
According to William's study, 
\textit{scholars}' papers are rather technical, jargonistic, specific and unappealing while \textit{influencers}' papers are rather non-technical, large-scale, pragmatic, relatable and appealing. \textit{Exceptionals}, on the other hand, combine the advantages of both sides.\footnote{For the full list of characterisics see \citet[p. 7]{Williams.2022}.}

Intuitively, we anticipate academic papers to be more likely to land in the category of \textit{scholars} as they, in theory, focus more on basic research that is more valuable to other researchers. Papers where the industry is involved, on the other hand, should in theory be more relevant for a wider audience,  represented by the category \textit{influencers}, since they traditionally perform more applied research. There should be no significant difference in the category \textit{exceptionals} if we assume similar levels of research quality across our groups. 

\subsection{Subfield Analysis}
For our subfield analysis, we match our given list of paper keywords with the controlled vocabulary of AI by 
\citet{NicolauDuranSilva.2021}. It describes an extensive and expert-validated list of terms for the following AI subdomains:
\begin{enumerate}
 \item General, 
 \item Machine Learning, 
 \item Computer Vision, 
 \item Natural Language Processing, 
 \item Knowledge Representation and Reasoning, 
 \item Distributed Artificial Intelligence, 
 \item Expert Systems, Problem-Solving, Control Methods and Search
 \item AI Ethics
\end{enumerate}
To analyze common research topics across groups, we measure the share of matched keywords in above mentioned subfields from company-involving papers.\footnote{Company-involving papers are the union of pure company and cooperation papers.} If this share is higher in a given subdomain than in the total average of all matched keywords, we consider that this subfield is of higher research focus for the industry.

\section{Results}
\label{sec:results}

Firstly, we present the results of our citation analysis of industry, academic, and academic-corporate collaboration papers in Sec.~\ref{subsec:citation}.
Secondly, we analyze the impact of AI papers across these groups based on altmetrics 
in Sec.~\ref{subsec:altmetric}.
Thirdly, we use the paper keywords to investigate the research focus of the three groups in Sec.~\ref{subsec:keyword}. 

\subsection{Citation Analysis}
\label{subsec:citation}

The bibliometric data in the MAG provides insights into the citation impact of published AI research. As stated in Section \ref{sec:data}, we start with around 230,000 papers, that we split into the affiliation groups of \textit{education}, \textit{company} and \textit{cooperation} according to their entry in GRID.

\subsubsection{Citation Statistics}
\label{subsubsec:statistics}

\begin{figure}[tb]
  \centering
  \includegraphics[width=.72\textwidth]{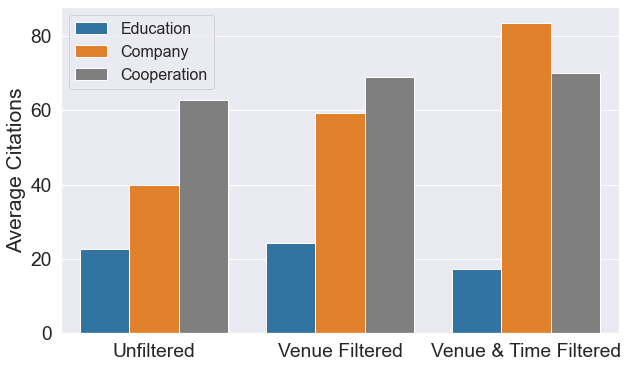}
  \captionof{figure}{Average citations for papers across groups}
  \label{fig:3}
\end{figure}

First of all, we investigate the effects the filters from data preparation step 4 have on citations across our three disjoint groups. In Fig.~\ref{fig:3} we directly compare the groups in terms of average citations per paper. We see that there are already  visible differences in average citations in the unfiltered sample. Cooperation papers have a clear lead in average citations. In fact, AI papers co-written by university and company departments earn over 50\% more citations as pure company papers and three times as many as pure university papers. 

The middle and right part of Fig.~\ref{fig:3} show the effect the two filters applied in preparation step 4 have on our groups. There, we can see that papers across the board have slightly higher average citations when only publications from journals and conferences are included. As predicted, this effect is even stronger for the average number of citations of company papers which feature a lot of patents in the unfiltered sample (59.3 vs. 40.0 citations on average). Their papers were cited almost 50\% more often on average after applying the venue filter.

Additionally filtering out papers written before 2012 reveals further noteworthy results. While citations in education-type papers fell according to our expectations, they increased considerably in company-type papers.
As expected, an AI research paper authored by an university after 2012 received less citations on average than an older paper. Corporate AI papers, on the other hand, received on average significantly more citations per paper after 2012. In fact, pure company papers reached an average of 83.6 citations per publication, which is almost 5 times the amount of citations an academic paper received. They even outperformed the cooperation papers, which follow close behind with 70 citations on average. 

\begin{figure}[tb]
  \centering
  \includegraphics[width=0.7\textwidth]{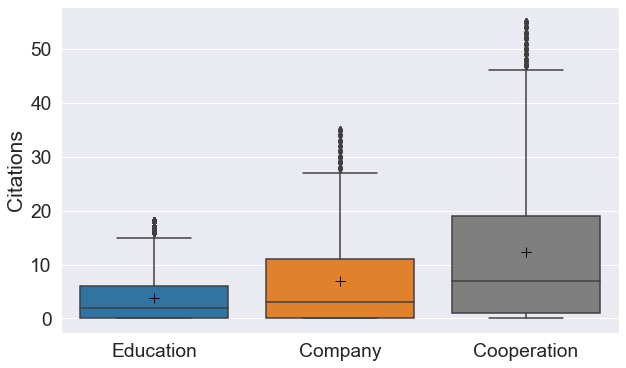}
  \captionof{figure}{Number of citations across groups without outlier papers.}
  \label{fig:4}
\end{figure}

\begin{figure}[tb]
  \centering
  \includegraphics[width=.78\textwidth]{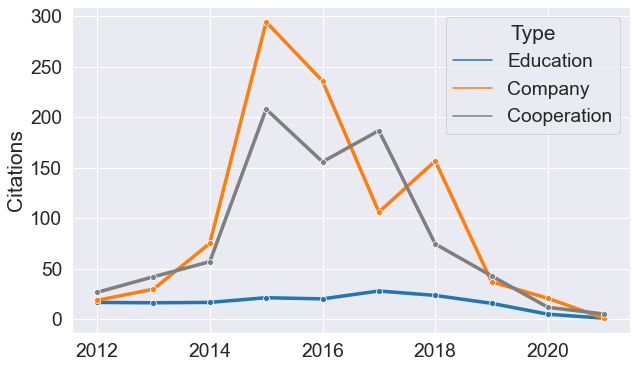}
  \captionof{figure}{Citation counts per paper across the groups from 2012-2021}
  \label{fig:5}
\end{figure}

Admittedly, the above mentioned differences in average citations are not representative for the whole citation distribution. Highly cited outlier papers distort these numbers in a major way. For example, a famous 2014 paper from the University of Toronto titled ``Dropout: a simple way to prevent neural networks from overfitting'' has been cited almost 20,000 times. In 2018, researchers from Google published a very successful AI paper named ``BERT: Pre-training of Deep Bidirectional Transformers for Language Understanding'' with more than 15,000 citations according to our dataset (which explains that year's slight bump in Fig.~\ref{fig:5}).

To evaluate the robustness of these findings, we for once exclude those outliers and visualize the distribution of the remaining data points across groups as boxplots. Fig.~\ref{fig:4} shows that the differences across groups still hold, even if such outlier papers are excluded. As we can see, most papers in all groups receive either zero citations or citations in the single digit area.  Nevertheless, company papers and cooperation papers have their boxes shifted upwards. Cooperation papers in particular show a substantially higher median count and upper quartile. This suggests that the previously described effects are notably robust and show up even when disregarding exceptionally successful outlier papers.

Moreover, we look into the average number of citations per year to capture time-dependent effects (see Fig.~\ref{fig:5}). We observe that in 2012 citations across the three groups were nearly identical. Company and cooperation papers only started having higher averages since 2013. The clear peak, however, is reached in 2015. There, papers from pure companies averaged about 300 citations and academic-industry collaboration papers 200 citations, which are outstanding numbers. Papers written entirely at universities at the same time receive only 21 citations. Overall, their mean citation counts remain quite stable around the mark of 20 throughout the years and thus always beneath their counterparts with industry involvement. As expected, citations drop for newer papers across the board and as a result the differences diminish over the last 3 years.

\subsubsection{Percentile Rank Results}
\label{subsubsec:prresults}

\begin{figure}[tb]
 \begin{minipage}[b]{0.3\linewidth}
  \centering
  \textbf{Bottom 50\%}\par\medskip
  \includegraphics[width=\textwidth]{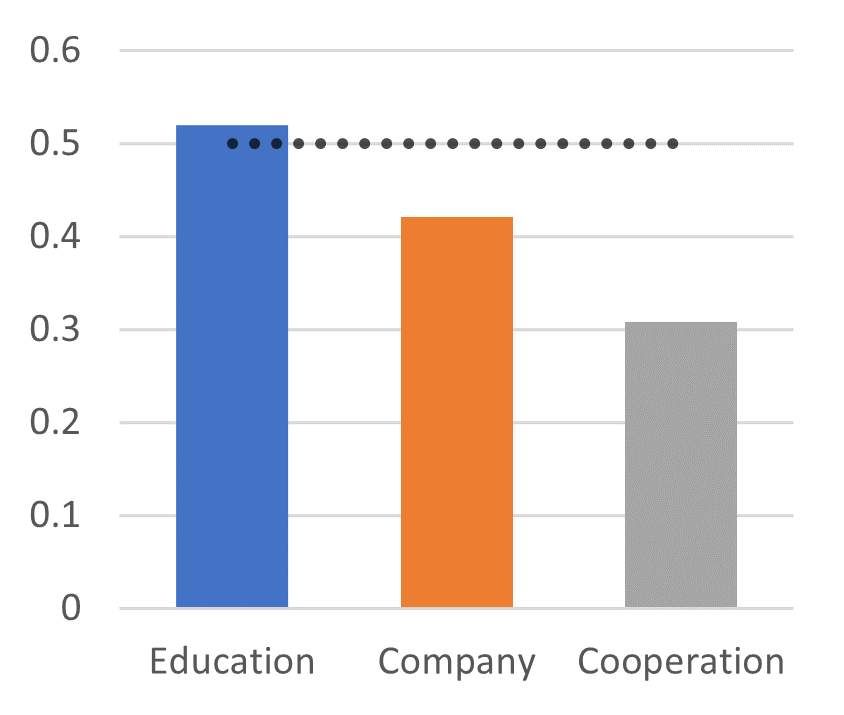} %
 \end{minipage}
 \hspace{0.01\linewidth}
 \begin{minipage}[b]{0.3\linewidth}
  \centering
  \textbf{50th-75th}\par\medskip
  \includegraphics[width=\textwidth]{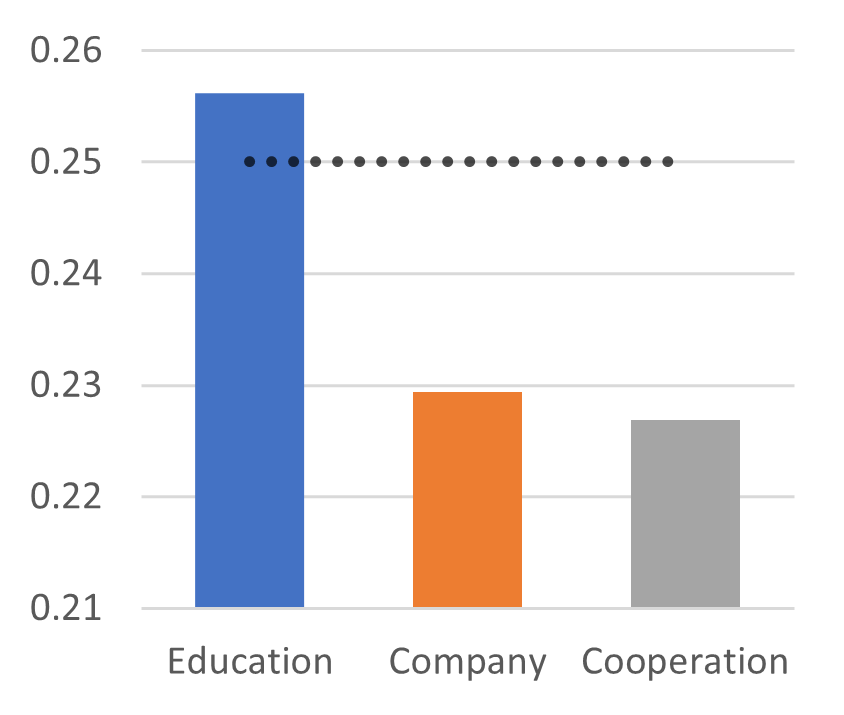}%
 \end{minipage}
 \hspace{0.01\linewidth}
 \begin{minipage}[b]{0.3\linewidth}
  \centering
  \textbf{75th-90th}\par\medskip
  \includegraphics[width=\textwidth]{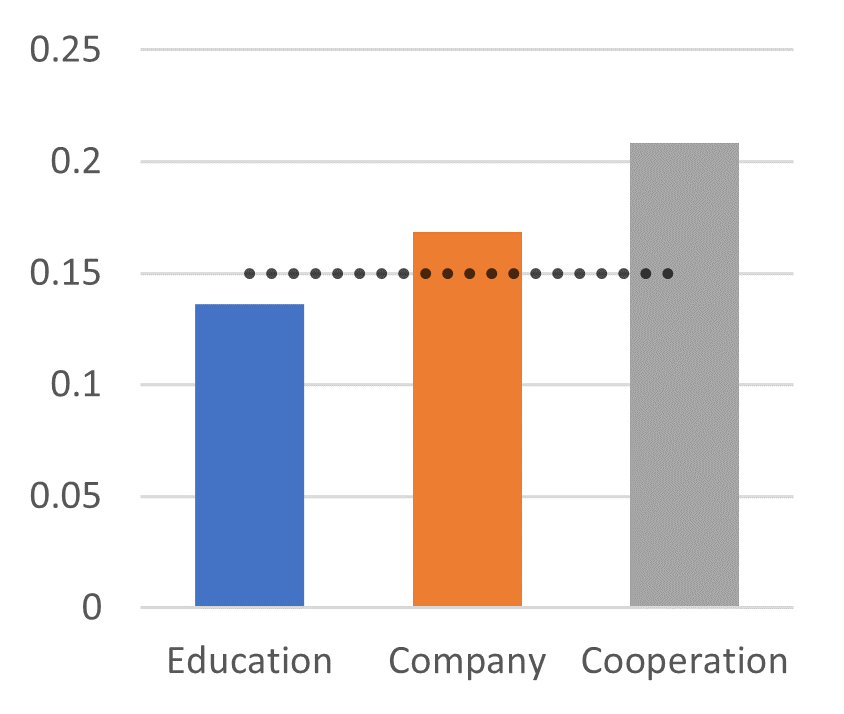}%
 \end{minipage}
 \newline
 \begin{minipage}[b]{0.3\linewidth}
  \centering
  \textbf{90th-95th}\par\medskip
  \includegraphics[width=\textwidth]{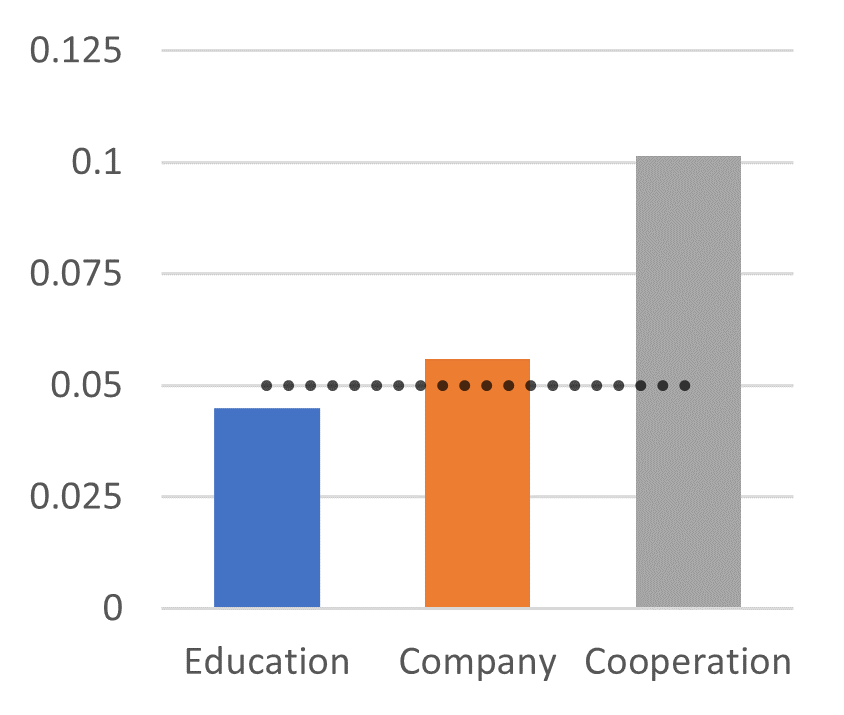} %
 \end{minipage}
 \hspace{0.01\linewidth}
 \begin{minipage}[b]{0.3\linewidth}
  \centering
  \textbf{95th-99th}\par\medskip
  \includegraphics[width=\textwidth]{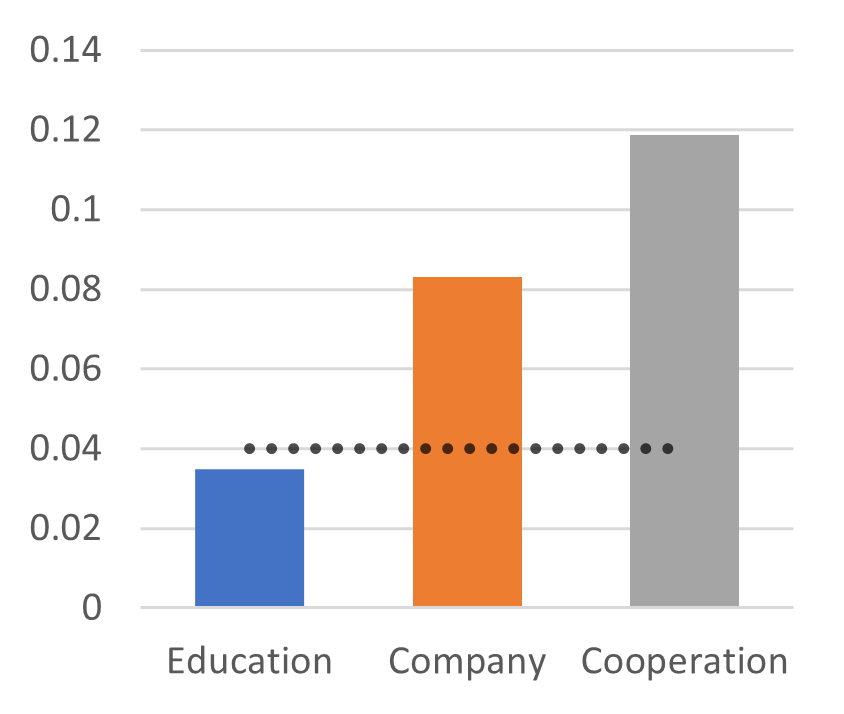} %
 \end{minipage}
 \hspace{0.01\linewidth}
 \begin{minipage}[b]{0.3\linewidth}
  \centering
  \textbf{Top 1\%}\par\medskip
  \includegraphics[width=\textwidth]{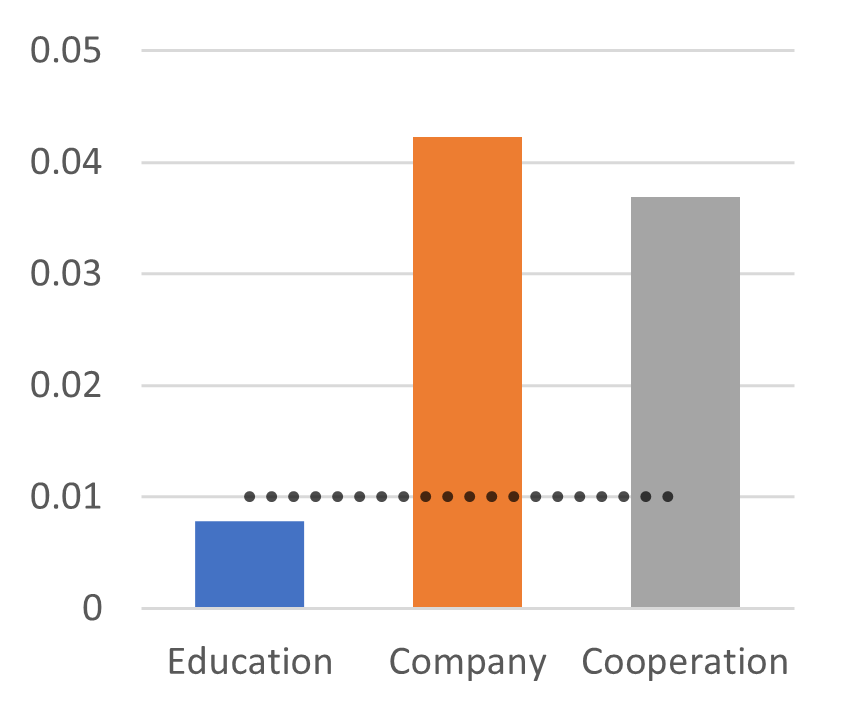} %
 \end{minipage}
 \captionof{figure}{Share of papers per group in the percentile rank classes in relation to the expected share}
 \label{fig:7}
\end{figure}

We also make use of the PR-6 approach to measure the impact of AI papers. 
We split the \textit{citation dataset} papers into the 6 percentile classes according to their standing in the ordered list of papers and evaluate the share of papers from each group that get assigned to the classes (see Fig.~\ref{fig:7}). 
The shares these charts depict can be interpreted as the probability that a random paper from a particular group will end up in a percentile class. The expected share of a randomly drawn paper from the aggregated sample is marked with a dotted reference line (e.g., 25\% of papers are expected to land in the 50th-75th percentile class and so on).
As we can see in the 'bottom 50\%' class, the share is higher for education-type papers, lower for company papers, and lowest for cooperation papers. This, however, only holds true for the first two classes, which contain papers with relatively low citation impact. From the 75th to 90th percentile class and onwards this tendency reverses. Papers written by universities are underrepresented in the upper classes and papers written by the industry or in academia-industry collaboration are clearly overrepresented. This effect increases the higher the percentiles get. In the top 1\% class, the share for company papers even exceeds that of cooperation papers. 

Furthermore, we calculate the weighted percentile rank score (see Sec.~\ref{sec:percentile-rank}) of each group given formula \eqref{formula1}. Since the education group makes up the vast majority of papers in our sample, it yields almost precisely the expected score of $1.91$ as in the formula \eqref{formula2}. Company papers, on the other hand, produce a score of $2.34$ and cooperation papers $2.67$.

To compare our results with other paper selection approaches from the literature \citep{Ahmed.22.10.2020,Hagendorff.2021,Hartmann.2020}, we prepare and analyze a smaller \textit{conference sample} with papers from 5 major AI conferences with our MAG data.
The resulting citation metrics are depicted in Table~\ref{tab:2}.
As can be seen, the averages and PR6-scores from industry-involving publications are comparable to the ones from our \textit{citation dataset}, except that the average and median of papers from academia are slightly higher. That latter fact is not surprising, since we expect papers that are accepted in the most prestigious conferences to be cited more frequently. However, this also shows that the inclusion of additional conference and journal papers in the \textit{citation dataset} results in a larger difference in citation metrics across the groups.

It follows that industry-involving publications in AI conferences have a higher citation impact than academic ones, but the magnitude of the effect is smaller than that of our \textit{citation dataset}, which also features journals and less prestigious conferences. %
To evaluate whether this tendency holds true for an unrelated field of study, we also applied it on our \textit{non-AI sample} consisting of electrical engineering papers.\footnote{The same data preparation steps, filters, and analysis methods are used as in our \textit{citation dataset}.} As the metrics in Table \ref{tab:3} show, we cannot observe a clear difference between papers from academia and the industry in citation counts for electrical engineering. Only, the cooperation papers seem to have a slight edge over the other groups.

Further, we statistically tested the significance of the observed citation differences across the groups. The null hypothesis of the Kruskal-Wallis, that there are no differences in our \textit{citation dataset}, does not hold (p-value $\ll 0.001$) and thus we continued with pairwise Mann-Whitney-U tests. They showed that both company and cooperation papers have significantly higher citations than education-type papers (p-value $\ll 0.001$). Remarkably, cooperation papers also show a significant difference to company papers (p-value $\ll 0.001$). Given the above, we can confidently say that the citation impact of academia-industry collaboration papers is higher than that of pure company papers, which in turn have a higher impact than purely academic ones.
The tests for our \textit{conference sample} are in line with the tests of our \textit{citation dataset} as they prove to be statistically significant as well. %

On the contrary, the Mann-Whitney-U tests of our \textit{non-AI sample} could not show significantly different citations numbers between education and company papers in either direction, hinting on the fact that the AI research field is unique in this regard.
There, only cooperation papers are significantly better in attracting citations than the other groups.
\begin{table}[tb]
 \centering
 \caption{Citation statistics of AI conference papers}
 \begin{tabular}{lrrrrrr} 
  \toprule
  \multicolumn{1}{l}{Group} & \multicolumn{1}{l}{Paper Count} & \multicolumn{1}{l}{Mean} & \multicolumn{1}{l}{Median} & \multicolumn{1}{l}{Max} & \multicolumn{1}{l}{Std} & \multicolumn{1}{l}{PR6} \\
  \cmidrule(lr){1-7}
Education   & 19.8k  & 24.26  & 5.00  &  42,481  & 349.06  & 1.88 \\
Company     &  1.2k  & 80.09  & 5.00  &  13,138  & 633.08  & 2.11 \\
Cooperation &  3.3k  & 37.60  & 8.00  &   2,819  & 141.23  & 2.21 \\
  \bottomrule
  \vspace{0.1cm}
 \end{tabular}
 \label{tab:2}
\end{table}

\begin{table}[tb]
 \centering
 \caption{Citation statistics of engineering papers}
 \begin{tabular}{lrrrrrr} 
  \toprule
  \multicolumn{1}{l}{Group} & \multicolumn{1}{l}{Paper Count} & \multicolumn{1}{l}{Mean} & \multicolumn{1}{l}{Median} & \multicolumn{1}{l}{Max} & \multicolumn{1}{l}{Std} & \multicolumn{1}{l}{PR6} \\
  \cmidrule(lr){1-7}
Education   & 110k  & ~8.34  & 2.00  & ~2,303 & ~30.88  & 1.98 \\
Company     &   6k  & 7.37  & 2.00  &   722 & 22.82  & 1.95 \\
Cooperation &   6k  & 9.94  & 3.00  &   427 & 20.43  & 2.24 \\
  \bottomrule
  \vspace{0.1cm}
 \end{tabular}
 \label{tab:3}
\end{table}

\subsubsection{Top Research Organizations}
\label{subsubsec:top}

As observed earlier, the citation distribution is highly skewed, with some papers receiving thousands of citations, while others do not get recognized at all. 
To identify which organizations impact the field the most,
we take a look at the upper end of the top research producing organizations by publication and citation count. 

\begin{figure}[tb]
 \centering
 \includegraphics[width=.66\textwidth]{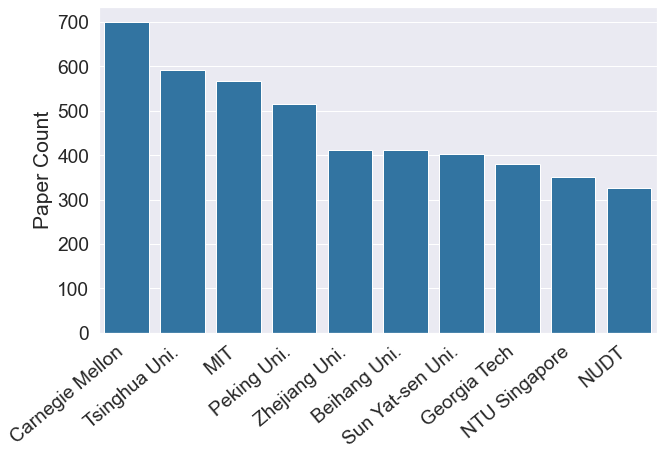} %
 \hspace{0.11cm}
 \centering
 \includegraphics[width=.66\textwidth]{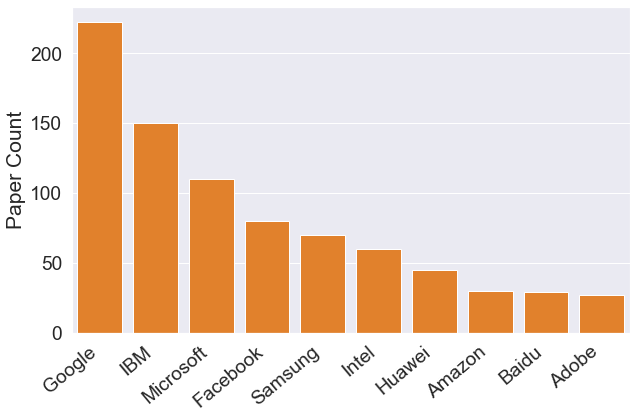} %
 \captionof{figure}{Top 10 paper counts of universities (top) and companies (bottom)}
 \label{fig:8}
\end{figure}

Starting with the contributions of the research institutions that were labeled as education types, we identified almost 4,000 
of them publishing in sum over 67,800 papers on the topic. Unsurprisingly, some academic research institutions publish significantly more AI articles than others. For instance, the Carnegie Mellon University, Tsinghua University, Massachusetts Institute of Technology, and Peking University are each responsible for over 500 articles, while there are many universities publishing only sporadically ($x_{0,5}= 4,$ $\overline{x}=17$). In Fig.~\ref{fig:8}, the top ten AI-article producing universities are depicted in descending order. Apart from the aforementioned four, the list follows with four Chinese universities, namely, Zhejiang University, Beihang University, Sun Yat-sen University, National University of Defense Technology, and has the Georgia Institute of Technology from the US and the Nanyang Technological University from Singapore in between. All of them have published 320 to 420 AI-articles respectively in the last decade. 

\begin{figure}[tb]
  \centering
  \includegraphics[width=.66\textwidth]{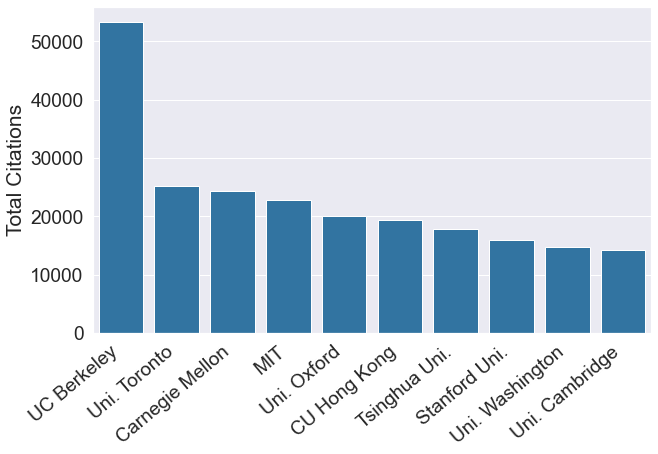} %
  \hspace{0.11cm}
  \centering
  \includegraphics[width=.66\textwidth]{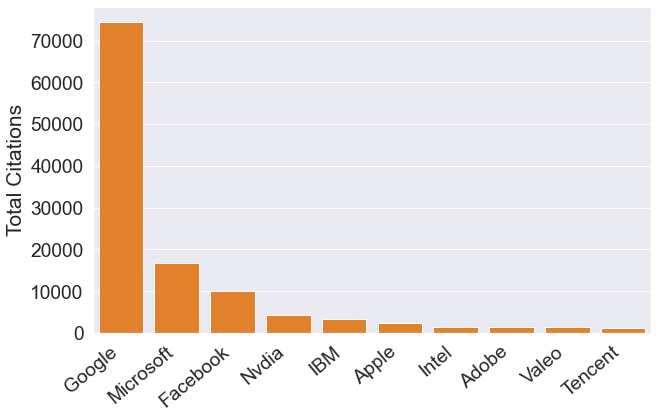} %
 \captionof{figure}{Top 10 total citations of universities (top) and companies (bottom)}
 \label{fig:9}
\end{figure}

The slope is even steeper for AI-papers authored by companies only. Fig.~\ref{fig:8} shows that even within the top ten of companies there are clear differences in publication counts. Google leads the board by a wide margin with 222 published papers followed by IBM with 150, Microsoft with 110, and Facebook with 80. The rest of the top ten features Samsung, Intel, Huawei, Amazon, Baidu and Adobe Systems\footnote{The mentioned company names are directly derived from the MAG and may not represent their correct legal name, e.g., 'Google' refers to its parent company Alphabet Inc. which also includes DeepMind, and 'Facebook' is now officially called Meta Platforms Inc., etc.} 
who are responsible for over 260 articles in sum. The majority of the other 160 companies included in the dataset have published significantly less AI research ($x_{0,5}= 2,$ $\overline{x}=9$). 

\begin{figure}[tb]
 \centering
 \includegraphics[width=0.55\textwidth]{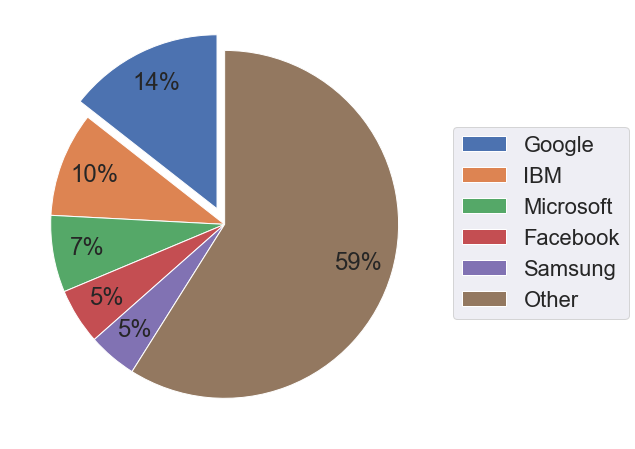} %
 \hspace{0.1cm}
 \centering
 \includegraphics[width=0.55\textwidth]{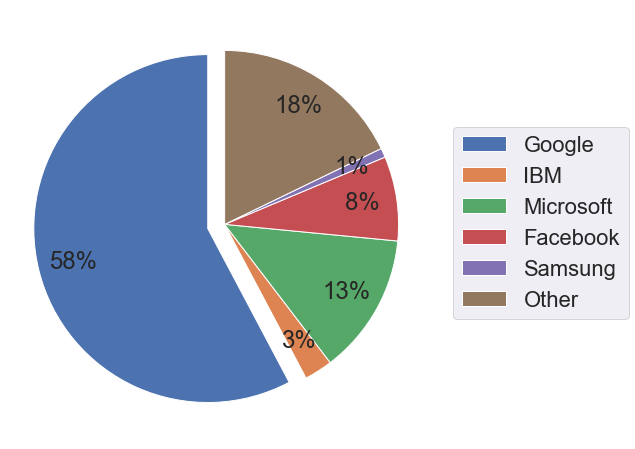} %
 \captionof{figure}{Share of papers (top) and citations (bottom) by companies}
 \label{fig:10}
\end{figure}

However, if we want to measure which research organizations have the greatest impact on the scientific community, we need to look at aggregated citations. Therefore, we selected the top ten academic and industry institutions in terms of citations and ranked them in descending order (see Fig.~\ref{fig:9}). On the academic side, we now have a clear winner in terms of the highest overall citation count of their AI papers: the University of California, Berkeley with over 53,000 citations. Most Chinese universities from the top ten by paper count have been replaced in the top ten by citation count either by other American universities, namely Stanford University and the University of Washington with around 16,000 and 15,000 citations respectively, or by the British universities of Oxford and Cambridge with around 20,000 and 14,000, respectively. The University of Toronto is also worth mentioning in second place with a total of over 25,000 citations. 

Google has by far the highest citation impact from the industry side, as can be seen in Fig.~\ref{fig:9}. In fact, with over 74,000 total citations, Google even surpasses the top universities from our list. Microsoft is in second place with nearly 17,000 citations, followed by Facebook/Meta with just under 10,000. Nvidia, IBM, and Apple papers have collected about 4,000, 3,000 and 2,000 citations, respectively, in that order. The top ten continues with Intel, Adobe Systems, Valeo and Tencent. Remarkably, the top eight of the list were all American tech companies, followed by a French automotive supplier and a Chinese technology holding company. 

Starting from all companies that have published AI papers, we looked at which companies are the most represented in terms of the number of publications and the number of citations of their publications (see Fig.~\ref{fig:10}). While Google researchers produced nearly 1 out of 7 of the industry's AI papers, they received the majority of the citations. Together with Microsoft and Facebook, the proportion of the total citations adds up to almost 80\%. This illustrates the high concentration of corporate research and its dependence on a few tech companies, as highlighted by \citet{Ahmed.22.10.2020}.

If we exclude papers authored by Google from our \textit{citation dataset}, we find that the citation average of company papers is more than halved ($x_{0,5}= 5;$ $\overline{x}=41.3$) compared to the results in Sec.~\ref{subsubsec:statistics}, which reinforces the great impact of their publications.
However, company papers still outperform university papers ($x_{0,5}= 3;$ $\overline{x}=17.23$), thus the observed difference does not solely rely on one company.
Even when excluding the top-3 AI-research producing companies, a considerable difference between company and education papers still persists, but admittedly on 
a smaller scale ($x_{0,5}= 4;$ $\overline{x}=24.3$).

\subsection{Altmetric Analysis}
\label{subsec:altmetric}

After mapping papers from our \textit{citation dataset} from the MAG to the Altmetric database we are able to retrieve around 13,000 papers with valid entries in both databases. In this section, we show the results of the altmetric analysis 
given this data. 

\begin{table}[tb]
 \centering
 \caption{Statistics of Attention Scores}
 \begin{tabular}{lrrrrrr} 
  \toprule
  \multicolumn{1}{l}{Group} & \multicolumn{1}{l}{Paper Count} & \multicolumn{1}{l}{Mean} & \multicolumn{1}{l}{Median} & \multicolumn{1}{l}{Max} & \multicolumn{1}{l}{Std} & \multicolumn{1}{l}{PR6} \\
  \cmidrule(lr){1-7}
Education   & 12,639  & ~11.97  & 3.00   &  3,362.28  & ~61.82   & 1.91 \\
Company     &   294  & 28.62  & 3.00   &  3,051.42  & 215.02  & 2.10 \\
Cooperation &   709  & 12.88  & 3.00   &  660.80   & 45.30   & 2.14 \\
  \bottomrule
 \end{tabular}
 \label{table1}
\end{table}

\subsubsection{Attention Score Statistics}
Starting with Altmetric.com's main metric, the \textit{(online-)attention score}, we look into similarities and differences in the \textit{altmetric dataset}.
The resulting descriptive statistics are summarized in Table \ref{table1}. All groups share the same median score of 3. On average on the other hand, education and cooperation papers receive a score of around 12. This indicates a heavily skewed distribution with some papers receiving really high attention while the majority receives little to no attention according to the Altmetric.com score, similar as with citation counts.
Company papers receive more than twice the attention score on average compared to papers from the other groups, with a mean score of 28.6. Notably, this exact number has to be taken with care, since the subset of company papers is less extensive and hence shows substantially more variance and standard deviation than the others. 

The difference between cooperation papers and education papers is not apparent when looking at the median or average scores. To more robustly compare the differences across groups, we make use of the PR-6 approach again and calculate the weighted percentile rank scores as given in Equation \eqref{formula1}. 
As the education group makes up the vast majority of papers, it yields almost precisely the expected score of $1.91$ as given in Equation~\eqref{formula2}. Company and cooperation papers on the other hand, produce scores of about $2.1$. This stems from the fact that most of these papers are to be found above the 50th percentile of attention scores.

Despite the difference in PR6 being seemingly small, the altmetric scores of the three groups still have statistically significant differences according to a Kruskal-Wallis test ($p \ll 0.01$).    
The pairwise results of Mann-Whitney-U tests indicate significantly higher scores of company papers compared to education papers ($p \ll 0.01$), significantly higher scores of cooperation papers to education papers ($p \ll 0.01$), and similar scores between company and cooperation papers ($p = 0.43$).

\subsubsection{Comparisons Across Metrics}
\label{subsubsec:correlation}

\begin{figure}[tb]
  \centering
  \includegraphics[width=.5\textwidth]{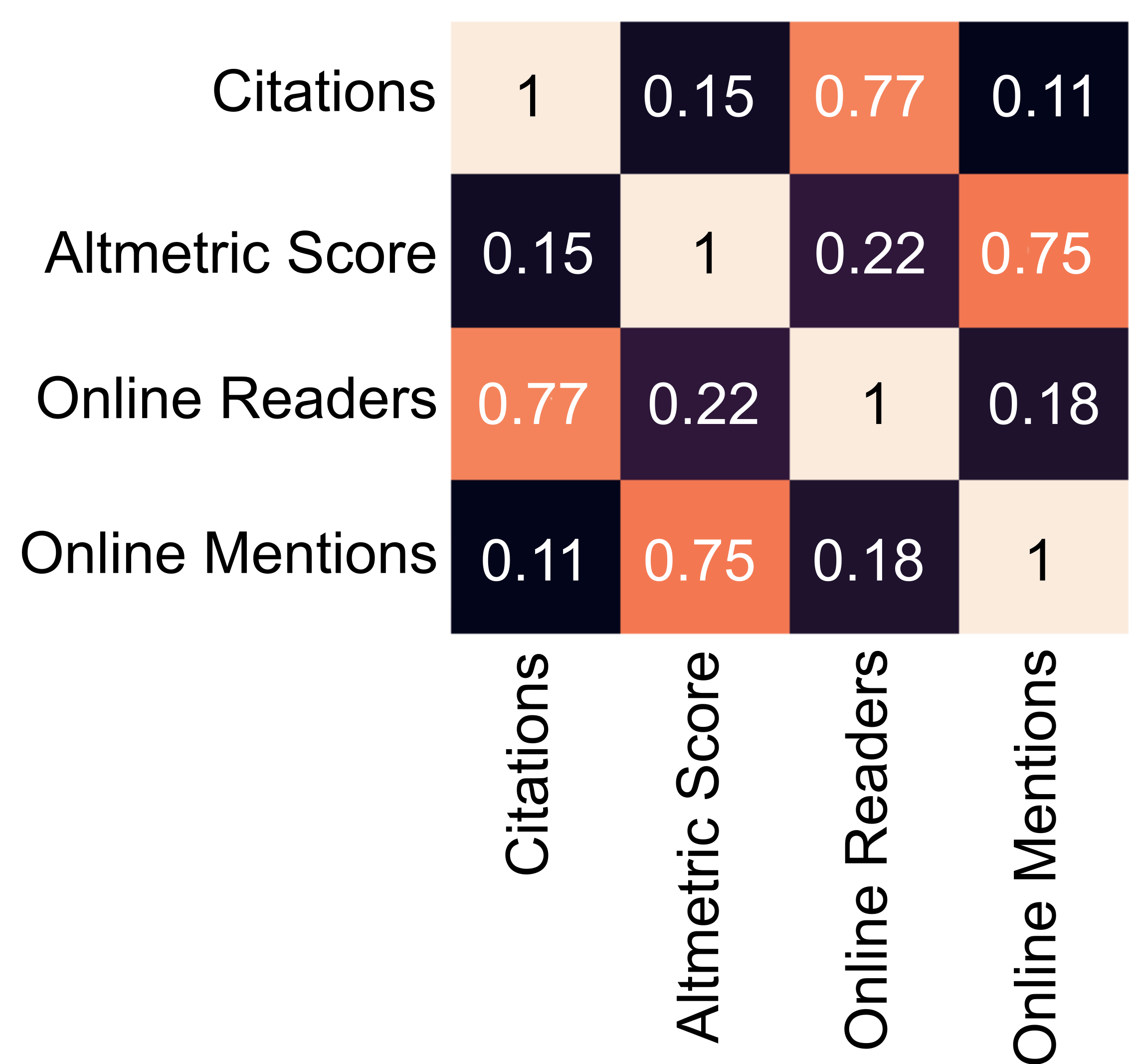} %
  \captionof{figure}{Pearson correlations between different metrics}
  \label{fig:11}
\end{figure}
In this subsection, we examine the relationship between altmetrics and citations. For this, we first look at the Pearson correlation coefficients between the metric scores, which are depicted as a matrix in Fig.~\ref{fig:11}.

As we can see, the assumption of an existing correlation between online mentions and the attention score is correct. 
Remarkably, however, the online reader counts are highly correlated to citations. Although not anticipated, it makes intuitively sense, since online reader counts are primarily derived from the Mendeley platform, which includes a reference management software that scientists use to organize related work. This suggests that the amount of researchers that have saved and read a paper on Mendeley is proportional to the citations it will later obtain, presumably because they kept the paper on Mendeley in order to later cite it. 
The citation counts and attention scores in the \textit{altmetric dataset} are only weakly correlated. This is in line with previous works that have compared altmetric scores to citations \citep{Salajegheh.2019,Zhang.2019}. 

\begin{figure}[tb]
  \centering
  \includegraphics[width=\textwidth]{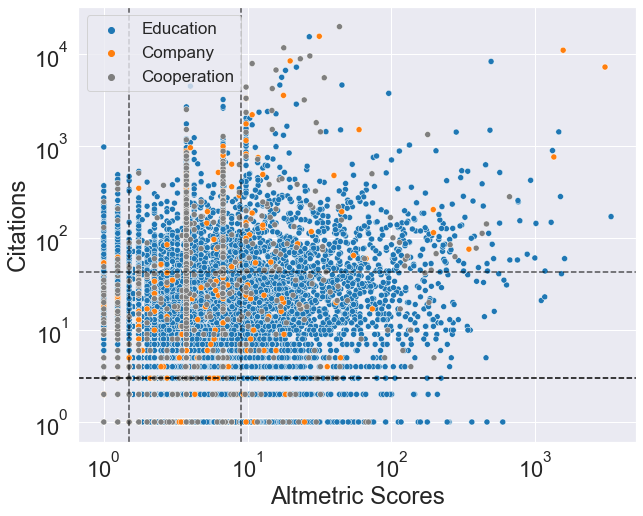} %
  \captionof{figure}{Citations counts vs. altmetric scores (in log scales)}
  \label{fig:12}
\end{figure}

In Fig.~\ref{fig:12}, we show the log scale of the citation count and the altmetric score of our 13,000 papers to further investigate this relationship. 
As explained in Sec.~\ref{sec:approach},  we apply the categorization of 
\citet{Williams.2022} into \textit{scholars}' papers that score high in citations but not in altmetrics, \textit{influencers}' papers which do the opposite, and \textit{exceptionals}' papers that score high in both. The categories are visualized in Fig.~\ref{fig:12} by the dashed lines: the upper left corner corresponds to \textit{scholars}' papers, the lower right corner to \textit{influencers}' papers and the upper right to the \textit{exceptionals}.
By doing so, we can see whether either of our groups performs better in attracting citations from the scientific world or attention on the web. 

Surprisingly, education, company, and cooperation papers are almost equally likely to be labeled as \textit{scholars}' or as \textit{influencers}' papers, which means that we cannot conclude a tendency of either group towards citations nor online attention (see Table~\ref{tab:4}). 
Nevertheless, there is a clear difference in the category of \textit{exceptionals}: Over 10\% of papers where companies were involved land in there in contrast to just under 6\% for papers with academic origins. In other words, our results show the relative superiority of industry papers over academic ones in both dimensions symmetrically. 

\begin{table}[tb]
 \centering
 \caption{Paper category shares per group type}
 \begin{tabular}{lrrr} 
  \toprule
  \multicolumn{1}{l}{Group} & \multicolumn{1}{l}{Scholar} & \multicolumn{1}{l}{Influencer} & \multicolumn{1}{l}{Exceptional}  \\
  \cmidrule(lr){1-4}
Education   & 0.02 & 0.03  & 0.06 \\
Company     & 0.01 & 0.02  & 0.11 \\
Cooperation & 0.02 & 0.03  & 0.10 \\
  \bottomrule
 \end{tabular}
 \label{tab:4}
\end{table}

\subsection{Keyword Analysis}
\label{subsec:keyword}

\begin{figure}[tb]
  \centering
  \includegraphics[width=\textwidth]{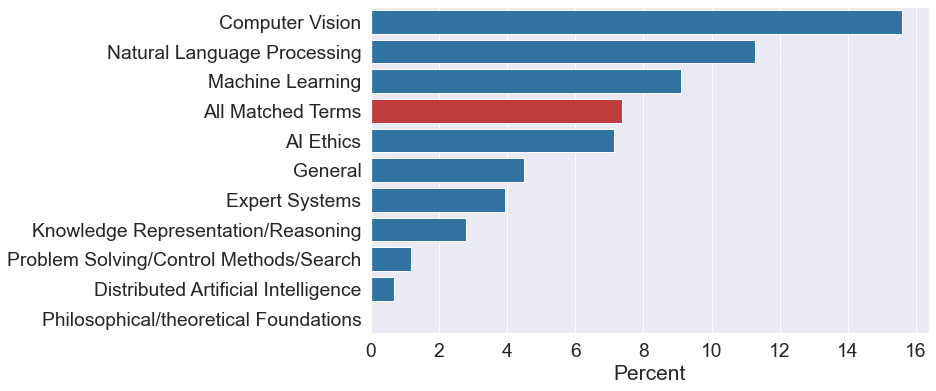} %
  \captionof{figure}{Percentage of matched terms attributable to papers with industry involvement}
  \label{fig:13}
\end{figure}

Our list of MAKG tags is derived from papers included in our \textit{citation dataset} presented in Sec.~\ref{subsec:citation}. It consists of about 310,000 tags, 93\% of which belong to pure university papers. Accordingly, the remaining 7\% of tags come from papers with company involvement. 
We match the list of tags with the terms from the controlled vocabulary for AI subfields \citep{NicolauDuranSilva.2021}. Again, about 7\% of all matched terms can be attributed to company or cooperation papers. 
However, the percentage of matched terms attributable to papers with company involvement differs across the AI subfields. Fig.~\ref{fig:13} illustrates this discrepancy.

The terms of the subfields 'Philosophical/theoretical Foundations', 'Distributed Artificial Intelligence' and 'Problem Solving, Control Methods and Search' have hardly any matches in company-involved papers. 
Terms of the subfields 'Knowledge Representation and Reasoning', 'Expert Systems' and 'General' receive some mentions in industry-involving papers but remain rare. This suggests that these six fields 
are not a research priority for the industry and are predominantly studied by academic institutions. 
In contrast, terms from the subfields 'Machine Learning', 'Natural Language Processing' and 'Computer Vision' appear more frequently in papers (co-)authored by companies. 'Computer Vision' in particular has more than twice the percentage compared to the average share of all matched terms. 
Therefore, these three fields can be considered more of a priority for companies. 
For 'AI Ethics' the percentage is very close to the average of 7\% and does not hint in any direction, which is consistent with the analysis of \citet{Hagendorff.2021}. With our approach, we can therefore not conclude whether AI ethics are more of a priority for academia or the industry.

\section{Discussion}
\label{sec:discussion}

We interpret and put our analysis results into context in Sec.~\ref{subsec:interpretation}, discuss potential explanations for our findings in Sec.~\ref{subsec:explanations}, and review possible implications for science, industry, and politics in Sec.~\ref{subsec:implications}. 

\subsection{Interpretation of Results} %
\label{subsec:interpretation}

Our citation analysis 
in Section \ref{subsec:citation} 
reveals several key points: 

First, papers involving companies in our main dataset clearly show higher 
mean, median, and PR-6 scores with respect to citations, 
indicating a higher citation impact of such publications. This finding is quite robust, as it is true even when very successful outlier papers or papers from top organizations are excluded. 
It provides additional support and reinforces the findings of 
\citet{Hagendorff.2021} and 
\citet{Klinger.22.09.2020}, who assert that there has been a discernible narrowing of AI research and a stagnation of diversity in recent years. Their research indicates that AI studies involving the private sector tend to exhibit lower levels of diversity and possess greater influence compared to academic research. 
One may assume that our findings contradict the statement of \citet[p. 367]{Hartmann.2020}, who argue that citation impact of academic and corporate research outputs is about the same in AI. However, since their study mainly considers bibliometric data from before 2012, their conclusion may very well be correct for the time frame under study (see our next finding).

The second finding is that the citation counts of AI research papers written by companies in the last decade are considerably higher than those written earlier. We observe this fact when applying the time filter and when analyzing the time series. This is remarkable considering that, on average, university AI papers follow the typical rule of lower citations for more recent publications. This might explain the roughly equal citation impact that \citet{Hartmann.2020} observed in their analysis of papers from 2004 to 2016. Moreover, our inspection of the top contributors supports the statement by 
\citet{Ahmed.22.10.2020} who proclaim that AI is increasingly being shaped by large technology companies. Google, Facebook/Meta, and Microsoft in particular stand out in our analysis as being highly influential. 

Third, cooperation papers have similarly high citation metric scores as pure company papers. They are slightly lower on average on our main sample but have in turn a higher median and PR-6 score. From that we can conclude that cooperation papers have about the same or even slightly higher citation impact than pure company papers.

However, we cannot directly conclude from the higher citation impact of papers that they are also of higher quality as the relationship between citation impact and research quality is complex. \citet{Aksnes.2019} describe research quality as a multidimensional concept that includes not only the aspect of scientific relevance -- for which citations are a good proxy for -- but also plausibility, originality, and societal relevance.

Including altmetric data analysis on the same papers 
allows us to also capture differences in societal relevance (or broader impact) of papers between groups, as argued by \citet{LutzBornmann.2014}. Our study in Sec.~\ref{subsec:altmetric} shows that citation counts and altmetric scores of papers correlate only weakly, suggesting that they indeed measure different things. We also observe that the high correlation found between citation counts and online readers suggests that this altmetric indicator might be a suitable feature for future citation predictions (see, e.g., \citet{Yan.2011,Yu.2014}), which we do not cover here.

Our analysis of the altmetric data provides additional evidence supporting the hypothesis of a higher impact of papers authored or co-authored by companies. Although the sample size used for this step was smaller and the sample contained relatively few industry papers, the observed differences remain both evident and statistically significant. Papers originating from the private sector not only garner significantly more citations but also attract greater online attention, implying a broader relevance beyond the realm of academia.

Hence, under the reasonable assumption that the plausibility and originality constituents of research quality are not considerably worse in corporate AI science,\footnote{The high acceptance rates of industry papers in leading AI journals (see, e.g., \citet{Ahmed.22.10.2020}) may rather indicate that they even outperform university papers in these regards, too.} our results indicate that papers with company-involvement are indeed of higher quality. As our comparison with papers from a non-AI field has shown, this tendency does not necessarily hold true for other branches of science, which supports the claim that the comparative research advantages of companies are only present in the field of AI \citep{Ahmed.22.10.2020}. 

Additionally, our keyword analysis in Sec.~\ref{subsec:keyword} showed that the industry predominantly focuses on the trending AI subdomains of computer vision, natural language processing, and machine learning, which are being impacted by advances in deep learning and benefit from it. On the contrary, academia has explored a wider variety of topics in AI, including knowledge representation and reasoning, distributed AI, and the theoretical foundations of AI.

\subsection{Potential Explanations}
\label{subsec:explanations}

Our analysis of corporate and university AI research over the last decade reveals a striking finding: corporate AI research has consistently produced papers with significantly higher impact than those from academia. In this subsection, we present three possible explanations for this observation.\\

\textit{Hypothesis 1: Industry's Access to Top AI Talent.}

One plausible explanation for the higher research impact of corporate AI papers is the growing trend of tech companies attracting top AI scientists from universities. Numerous studies \citep{Hagendorff.2021,Hartmann.2020,Jurowetzki.2021} have pointed out that the industry actively recruits and hires talented researchers, resulting in a considerable pool of AI expertise transitioning to corporate settings. While it may be tempting to attribute the entire increase in research impact to this talent migration, \citet{Jurowetzki.2021} found that AI scientists transitioning into industry experience an \textit{initial} surge in paper citations, indicating that the private sector possesses inherent advantages in conducting impactful AI research. Nevertheless, the precise extent to which the strategic hiring of successful university researchers contributes to the higher impact of corporate AI research remains an open question \citep{Hagendorff.2021}.\\

\textit{Hypothesis 2: Industry's Superior Computational Resources.}

\citet{Ahmed.22.10.2020} propose another explanation, suggesting that the unequal access to computational resources could be a key factor behind the prevalence of companies in AI research. Given that modern AI research heavily relies on extensive computing power (see deep learning and especially fields such as large language model training and meta-learning), large tech companies equipped with advanced hardware resources may be better positioned to make significant progress in computationally intensive research. \\

\textit{Hypothesis 3: Industry's Exclusive Data Assets.}

It is widely recognized that the private sector has gathered substantial quantities of proprietary data for AI research in recent years. According to \citet{Hartmann.2020}, this privileged access to data sets provides companies with a competitive edge over academic institutions in the field of AI research. Unfortunately, universities frequently lack access to such proprietary data, which hampers their capacity to conduct research that is on par with industry standards. 
The authors also propose that this data exclusivity may facilitate companies in appropriating the value of their research, making it easier for them to publish their results.\\

Notably, the last two explanations hold particular relevance for research on topics related to deep learning, as these methods require large amounts of data and considerable computational resources. As we saw in Sec.~\ref{subsec:keyword}, the industry shows a particular interest in areas most affected by deep learning. This finding supports the notion that the availability of both data and computational power contributes to the higher impact of company-involved AI research.

\subsection{Recommendations for Action for Academia, Industry, and Politics} %
\label{subsec:implications}

It is widely acknowledged that companies play a crucial role in driving AI research forward. They make significant contributions to the field by expanding research opportunities and facilitating evaluations in specific domains. Notable breakthroughs in AI research, such as advancements in medicine (e.g., drug discovery), chemistry (e.g., protein folding \citep{Jumper.2021}), and personal assistance systems (e.g., Alexa, ChatGPT), have been made possible through the support and funding provided by companies.

While the importance of industry involvement in AI research has been recognized in academic works \citep{Gil.07.08.2019}, and potential challenges associated with increased industry funding and participation have been noted \citep{Abdalla.2021,Irani.2019}, there has been limited systematic assessment on whether the specific characteristics of companies, such as their commercial orientation, impose undue limitations or have unforeseen impacts on AI research \citep{Littman.2021}. The question remains whether action needs to be taken by academia, industry, or policymakers to ensure a balance between academia and industry.

In the following, we propose concrete recommendations for action. They are based on the three hypotheses outlined in the previous subsection.\\

 \textit{Addressing Hypothesis 1: Bridging the AI Talent Gap}
\begin{enumerate}
    \item \textit{AI Research Fellowships and Career Paths:} Government and otherwise institutional fellowship programs and research grants may be established to attract and retain AI talent in academia, especially in job-wise competitive fields such as deep learning \citep{Jurowetzki.2021}. These programs should offer competitive funding packages and career development opportunities to motivate researchers to pursue academic careers  \citep{Jurowetzki.2021}. 
    Research indicates that a postdoc's transition to a non-academic career can be influenced various factors, including individual traits, the attitude and support of their principal investigator, as well as broader organizational and policy factors \citep{hayter2019factors}. In Germany, junior professorships were introduced in 2002 to enhance career prospects for young scientists. Nevertheless, postdocs reported only a marginal impact of these university reforms \citep{fitzenberger2014up}. %
 \item \textit{Industry-Academia Talent Development:} Developing joint training programs, internships, and mentorship initiatives between academia and industry can enable knowledge transfer between universities and industry, attract researchers to universities, and facilitate the sharing of cutting-edge research practices, data, and infrastructure \citep{spicer2022mind}.
\end{enumerate}

 \textit{Addressing Hypothesis 2: Providing Computational Resources for Academia} %
\begin{enumerate}
 \item \textit{Adequate Public Funding of Academic Computing Infrastructure:} 
 We argue that academia needs to be able to produce AI research of equal quality without depending on industry involvement, otherwise academic integrity may be compromised \citep{Jurowetzki.2021,Hagendorff.2021}. As such, it needs to be ensured that research institutions like universities have the necessary resources to conduct high-quality research independent of industry. 
 \item \textit{Provisioning Companies' Computing Infrastructure for Academia:} 
An alternative approach to enhancing academic research could involve providing cloud services and other infrastructure resources, such as high-performance computing, to academic researchers without imposing substantial additional costs or effort \citep{Littman.2021}. By doing so, this approach would address the hardware limitations currently faced by academia, fostering an environment conducive to extensive experimentation and innovation. One way to achieve this could be through partnerships between academic institutions and industry. These partnerships can facilitate joint research projects and knowledge sharing, while maintaining the integrity of academic research. Additionally, governments could subsidize companies that provide computing infrastructure specifically for AI research. 
\end{enumerate}
 \textit{Addressing Hypothesis 3: Promoting Openess 
 and Industry-Academia Collaborations} %
 
\begin{enumerate}
 \item \textit{Regulations and Incentives for Sharing Research Artifacts:} %
 It is essential that research results can be replicated and that artifacts, such as datasets, are accessible following the FAIR principles \citep{wilkinson2016fair}. 
 While the scientific community increasingly embraces the practice of sharing code and data,\footnote{See initiatives like the German National Research Data Infrastructure, \url{https://www.nfdi.de/}.} this culture has not yet become prevalent in the industry \citep{Ahmed.22.10.2020}. Instead, policies promoting transparency and the sharing of core datasets between public and private actors are subject to controversial discussions \citep{Cockburn.2018}. 
 Moreover, even if such resources are provided, the replication of experiments may be challenging due to limited hardware resources. 
In light of these challenges, it is important to facilitate the sharing of non-sensitive AI artifacts, such as datasets, code, and AI models, by private entities, while simultaneously ensuring compliance with privacy regulations.\footnote{Recent regulatory efforts being negotiated in the EU, namely the AI Act \& Data Act, point in the right direction and should further promote the disclosure of (research) data.} Encouraging incentives, such as tax benefits or research grants, should also be considered for organizations that openly share their artifacts. 
These measures would promote a culture of openness and collaboration, benefiting both the scientific community and industrial applications of AI technology. 
 \item \textit{Collaboration and Data Provisioning Platforms:} 
 It is highly beneficial to enhance and expand collaboration platforms and repositories that enable seamless sharing of AI code, models, and datasets (see, for instance, Zenodo, figshare, and AWS Open Data). By leveraging these platforms, researchers can effortlessly build upon one another's work, including works from industry, and replicate experiments, fostering a more cohesive and progressive research ecosystem. Zenodo and figshare in particular allow researchers to upload data at a mostly reasonable scale, but making very large datasets available is not possible on these platforms. In these cases, researchers currently rely primarily on corporate programs (e.g., Amazon's AWS Open Data). We recommend providing advanced data hosting options that are independent of companies. %
 \item \textit{Standardization of Data Formats and Documentation:} This facet involves advocating for the adoption of standardized data formats and documentation protocols. Such standardization enhances data interoperability and accessibility, thereby enabling researchers across institutions and industries to use common datasets with greater efficiency.

\item \textit{Interdisciplinary and Transdisciplinary Research Collaborations:} Encouraging collaboration between academia, industry, policymakers, and civil society organizations can foster (a) a unified perspective on the broad impact of AI research, and (b) more targeted strategies to address AI research challenges. Such collaboration can cultivate a holistic view, focusing on the development of value-oriented, long-term oriented, trustworthy AI. For instance, this approach can mitigate the systematic ``narrowing'' of AI research, where certain topics disproportionately receive more attention \citep{Klinger.22.09.2020, Hooker.14.09.2020, Jurowetzki.2021}. It is also beneficial for creating AI for social good, for instance when it comes to preserving diversity in AI models \citep{Kuhlman.27.02.2020} or combating disinformation \citep{Gil.07.08.2019, Buchanan.2021}. 
\item \textit{Incentives for Responsible AI Research Practices and Evaluation:} Funding mechanisms can be developed to reward researchers and organizations that actively incorporate ethical considerations into their AI research. This can be achieved through funding programs focusing on or prioritizing projects that are consistent with ethical guidelines \citep{AIethicsguidelines} 
such as the \textit{EU Ethics Guidelines for Trustworthy AI} (e.g., covering human agency, transparency, and fairness) \citep{EUethics,smuha2019eu} and the \textit{UN sustainability goals} \citep{vinuesa2020role}. 
In addition, policymakers could advocate for the widespread adoption of evaluation frameworks for researchers and institutions that do not rely solely on the citation count as impact indicator but that also encompass broader aspects, such as the openness and sustainability of research artifacts, to provide a more comprehensive and responsible measure of scientific contributions.  

\item \textit{Teaching Fair AI Research:} %
We can support the recommendations to provide mandatory courses or course content in schools and universities that provide students with the knowledge to be aware of industry influence on AI research and to reduce the bias of industry-driven AI research detached from corporate interests during the research processes. 
\end{enumerate}

\section{Conclusion}
\label{sec:conclusion}

Artificial Intelligence (AI) is one of the most significant technologies of our time. Thus, in various contexts, it becomes vital to understand and quantitatively measure who is exerting the greatest influence on the future of AI. In pursuit of this understanding, we used scientometric data from multiple academic databases to examine and compare the citations and altmetric influence of academic and corporate AI research papers. Although the vast majority of publications are still authored by academics, we found that the citation impact of a paper is significantly higher when it is (co-)authored by a company, confirming previous studies. Similarly, papers (co-)authored by companies receive significantly more attention online, as measured by altmetrics. The robustness of our results across different methods of data selection and metrics indicates that corporate AI research has indeed become more important than purely academic AI research in recent years.

We also provided an overview of the major players in corporate AI research. The publications of these key technology companies, mainly based in the United States and China, have an immense impact on the field. Additionally, our keyword analysis disclosed that the private sector is particularly interested in topics affected by deep learning, such as computer vision and natural language processing.

Based on our findings, we formulated several recommendations for action for academia, industry, and policymakers: 
(1) promoting ethical and responsible AI research,
(2) promoting open data and open code initiatives, the FAIR principles, and democratization of AI research,
(3) strengthening financial and infrastructural support for academic AI research, and
(4) bridging the AI talent gap. 
Given these recommendations, we are confident that corporate involvement can positively influence future AI research.

\section*{Declarations}

\textbf{Funding.} The authors did not receive support from any organization for the submitted work.

\noindent
\textbf{Competing interests.} The authors have no competing interests to declare that are relevant to the content of this article.

\bibliographystyle{spbasic_mod}  %
\bibliography{bibliography} %

\begin{thebibliography}{50}
\providecommand{\natexlab}[1]{#1}
\providecommand{\url}[1]{{#1}}
\providecommand{\urlprefix}{URL }
\expandafter\ifx\csname urlstyle\endcsname\relax
  \providecommand{\doi}[1]{DOI~\discretionary{}{}{}#1}\else
  \providecommand{\doi}{DOI~\discretionary{}{}{}\begingroup
  \urlstyle{rm}\Url}\fi
\providecommand{\eprint}[2][]{\url{#2}}

\bibitem[{Abdalla and Abdalla(2021)}]{Abdalla.2021}
Abdalla M, Abdalla M (2021) {The Grey Hoodie Project: Big Tobacco, Big Tech,
  and the Threat on Academic Integrity}. In: Proceedings of the 2021 AAAI/ACM
  Conference on AI, Ethics, and Society, New York, NY, USA, p 287–297,
  \urlprefix\url{https://doi.org/10.1145/3461702.3462563}

\bibitem[{Ahmed and Wahed(2020)}]{Ahmed.22.10.2020}
Ahmed N, Wahed M (2020) {The De-democratization of {AI:} Deep Learning and the
  Compute Divide in Artificial Intelligence Research}. \emph{CoRR}
  abs/2010.15581, \urlprefix\url{https://arxiv.org/abs/2010.15581}

\bibitem[{Aksnes et~al.(2019)Aksnes, Langfeldt, and Wouters}]{Aksnes.2019}
Aksnes DW, Langfeldt L, Wouters P (2019) {Citations, Citation Indicators, and
  Research Quality: An Overview of Basic Concepts and Theories}. \emph{SAGE
  Open} 9(1):2158--2440,
  \urlprefix\url{https://doi.org/10.1177/2158244019829575}

\bibitem[{Alom et~al.(2018)Alom, Taha, Yakopcic, Westberg, Sidike, Nasrin,
  Essen, Awwal, and Asari}]{Alom.03.03.2018}
Alom MZ, Taha TM, Yakopcic C, Westberg S, Sidike P, Nasrin MS, Essen BCV, Awwal
  AAS, Asari VK (2018) {The History Began from AlexNet: {A} Comprehensive
  Survey on Deep Learning Approaches}. \emph{CoRR} abs/1803.01164,
  \urlprefix\url{http://arxiv.org/abs/1803.01164}

\bibitem[{Bornmann(2014)}]{LutzBornmann.2014}
Bornmann L (2014) Do altmetrics point to the broader impact of research? an
  overview of benefits and disadvantages of altmetrics. \emph{Journal of
  Informetrics} 8(4):895--903,
  \urlprefix\url{https://doi.org/10.1016/j.joi.2014.09.005}

\bibitem[{Bornmann et~al.(2013)Bornmann, Leydesdorff, and Mutz}]{Bornmann.2013}
Bornmann L, Leydesdorff L, Mutz R (2013) The use of percentiles and percentile
  rank classes in the analysis of bibliometric data: Opportunities and limits.
  \emph{Journal of Informetrics} 7(1):158--165,
  \urlprefix\url{https://doi.org/10.1016/j.joi.2012.10.001}

\bibitem[{Buchanan et~al.(2021)Buchanan, Lohn, Musser, and
  Sedova}]{Buchanan.2021}
Buchanan B, Lohn A, Musser M, Sedova K (2021) {Truth, Lies, and Automation: How
  Language Models Could Change Disinformation}.
  \urlprefix\url{https://doi.org/10.51593/2021CA003}

\bibitem[{Cockburn et~al.(2018)Cockburn, Henderson, and Stern}]{Cockburn.2018}
Cockburn I, Henderson R, Stern S (2018) {The Impact of Artificial Intelligence
  on Innovation: An Exploratory Analysis}. In: The Economics of Artificial
  Intelligence, {National Bureau of Economic Research}, Cambridge, MA,
  \urlprefix\url{https://doi.org/10.3386/w24449}

\bibitem[{Dernis et~al.(2019)Dernis, Squicciarini, Nakazato, and {et
  al.}}]{Dernis.24.09.2019}
Dernis H, Squicciarini M, Nakazato S, {et al} (2019) \emph{{World Corporate Top
  R{\&}D investors: Shaping the Future of Technologies and of AI}}.
  \urlprefix\url{https://op.europa.eu/en/publication-detail/-/publication/c56f25b8-df3f-11e9-9c4e-01aa75ed71a1/language-en}

\bibitem[{Duran-Silva et~al.(2021)Duran-Silva, Fuster, Massucci, Parra-Rojas,
  Quinquill{\`a}, Roda, and Rondelli}]{NicolauDuranSilva.2021}
Duran-Silva N, Fuster E, Massucci FA, Parra-Rojas C, Quinquill{\`a} A, Roda F,
  Rondelli B (2021) A controlled vocabulary for research and innovation in the
  field of artificial intelligence (ai).
  \urlprefix\url{https://doi.org/10.5281/zenodo.4536033}

\bibitem[{{European Commission}(2019)}]{EUethics}
{European Commission} (2019) {COM/2019/168 final}.
  \urlprefix\url{https://ec.europa.eu/transparency/documents-register/api/files/COM(2019)168_0/de00000000080551?rendition=false}

\bibitem[{F{\"a}rber(2019)}]{Farber.2019}
F{\"a}rber M (2019) {{The Microsoft Academic Knowledge Graph: A Linked Data
  Source with 8 Billion Triples of Scholarly Data}}. In: Proceedings of the
  18th International Semantic Web Conference, ISWC'19, pp 113--129,
  \urlprefix\url{https://doi.org/10.1007/978-3-030-30796-7\_8}

\bibitem[{F{\"a}rber and Ao(2022)}]{Faerber.2022}
F{\"a}rber M, Ao L (2022) {The Microsoft Academic Knowledge Graph enhanced:
  Author name disambiguation, publication classification, and embeddings}.
  \emph{Quantitative Science Studies} 3(1):51--98,
  \urlprefix\url{https://doi.org/10.1162/qss_a_00183}

\bibitem[{Fitzenberger and Schulze(2014)}]{fitzenberger2014up}
Fitzenberger B, Schulze U (2014) {Up or Out: Research Incentives and Career
  Prospects of Postdocs in Germany}. \emph{German Economic Review}
  15(2):287--328, \urlprefix\url{https://doi.org/10.1111/geer.12010}

\bibitem[{Gil and Selman(2019)}]{Gil.07.08.2019}
Gil Y, Selman B (2019) A 20-year community roadmap for artificial intelligence
  research in the {US}. \emph{CoRR} abs/1908.02624,
  \urlprefix\url{http://arxiv.org/abs/1908.02624}

\bibitem[{Group et~al.(2019)Group, Irani, Salehi, Pal, Monroy-Hern\'{a}ndez,
  Churchill, and Narayan}]{Irani.2019}
Group CPS, Irani L, Salehi N, Pal J, Monroy-Hern\'{a}ndez A, Churchill E,
  Narayan S (2019) {Patron or Poison? Industry Funding of HCI Research}. In:
  Proceedings of the 22nd ACM Conference on Computer-Supported Cooperative Work
  and Social Computing, CSCW'19, p 111–115,
  \urlprefix\url{https://doi.org/10.1145/3311957.3358610}

\bibitem[{Hagendorff and Meding(2021)}]{Hagendorff.2021}
Hagendorff T, Meding K (2021) Ethical considerations and statistical analysis
  of industry involvement in machine learning research. \emph{AI {\&} Society}
  \urlprefix\url{https://doi.org/10.1007/s00146-021-01284-z}

\bibitem[{Hartmann and Henkel(2020)}]{Hartmann.2020}
Hartmann P, Henkel J (2020) {The Rise of Corporate Science in AI: Data as a
  Strategic Resource}. \emph{Academy of Management Discoveries}
  \urlprefix\url{https://doi.org/10.5465/amd.2019.0043}

\bibitem[{Harzing and Alakangas(2017)}]{Harzing.2017}
Harzing AW, Alakangas S (2017) {Microsoft Academic: Is the phoenix getting
  wings?} \emph{Scientometrics} 110(1):371--383,
  \urlprefix\url{https://doi.org/10.1007/s11192-016-2185-x}

\bibitem[{Hayter and Parker(2019)}]{hayter2019factors}
Hayter CS, Parker MA (2019) Factors that influence the transition of university
  postdocs to non-academic scientific careers: An exploratory study.
  \emph{Research Policy} 48(3):556--570,
  \urlprefix\url{https://doi.org/10.1016/j.respol.2018.09.009}

\bibitem[{Hooker(2021)}]{Hooker.14.09.2020}
Hooker S (2021) The hardware lottery. \emph{Commun {ACM}} 64(12):58--65,
  \urlprefix\url{https://doi.org/10.1145/3467017}

\bibitem[{Hug et~al.(2017)Hug, Ochsner, and Br{\"a}ndle}]{Hug.2017}
Hug SE, Ochsner M, Br{\"a}ndle MP (2017) {Citation analysis with Microsoft
  Academic}. \emph{Scientometrics} 111(1):371--378,
  \urlprefix\url{https://doi.org/10.1007/s11192-017-2247-8}

\bibitem[{Jobin and Ienca(2019)}]{AIethicsguidelines}
Jobin A, Ienca E Marcello an~Vayena (2019) {The global landscape of AI ethics
  guidelines}. \emph{Nature Machine Intelligence} p 389–399,
  \urlprefix\url{https://doi.org/10.1038/s42256-019-0088-2}

\bibitem[{Jumper et~al.(2021)Jumper, Evans, Pritzel, Green, Figurnov,
  Ronneberger, Tunyasuvunakool, Bates, {\v{Z}}{\'i}dek, Potapenko, Bridgland,
  Meyer, Kohl, Ballard, Cowie, Romera-Paredes, Nikolov, Jain, Adler, Back,
  Petersen, Reiman, Clancy, Zielinski, Steinegger, Pacholska, Berghammer,
  Bodenstein, Silver, Vinyals, Senior, Kavukcuoglu, Kohli, and
  Hassabis}]{Jumper.2021}
Jumper J, Evans R, Pritzel A, Green T, Figurnov M, Ronneberger O,
  Tunyasuvunakool K, Bates R, {\v{Z}}{\'i}dek A, Potapenko A, Bridgland A,
  Meyer C, Kohl SAA, Ballard AJ, Cowie A, Romera-Paredes B, Nikolov S, Jain R,
  Adler J, Back T, Petersen S, Reiman D, Clancy E, Zielinski M, Steinegger M,
  Pacholska M, Berghammer T, Bodenstein S, Silver D, Vinyals O, Senior AW,
  Kavukcuoglu K, Kohli P, Hassabis D (2021) {Highly accurate protein structure
  prediction with AlphaFold}. \emph{Nature} 596(7873):583--589,
  \urlprefix\url{https://doi.org/10.1038/s41586-021-03819-2}

\bibitem[{Jurowetzki et~al.(2021)Jurowetzki, Hain, Mateos{-}Garcia, and
  Stathoulopoulos}]{Jurowetzki.2021}
Jurowetzki R, Hain DS, Mateos{-}Garcia J, Stathoulopoulos K (2021) {The
  Privatization of {AI} Research(-ers): Causes and Potential Consequences -
  From university-industry interaction to public research brain-drain?}
  \emph{CoRR} abs/2102.01648, \urlprefix\url{https://arxiv.org/abs/2102.01648}

\bibitem[{Klinger et~al.(2020)Klinger, Mateos{-}Garcia, and
  Stathoulopoulos}]{Klinger.22.09.2020}
Klinger J, Mateos{-}Garcia J, Stathoulopoulos K (2020) A narrowing of {AI}
  research? \emph{CoRR} abs/2009.10385,
  \urlprefix\url{https://arxiv.org/abs/2009.10385}

\bibitem[{Krieger et~al.(2021)Krieger, Pellens, Blind, Gruber, and
  Schubert}]{Krieger.2021}
Krieger B, Pellens M, Blind K, Gruber S, Schubert T (2021) {Are firms
  withdrawing from basic research? An analysis of firmlevel publication
  behaviour in German}. \emph{Scientometrics} (126):9677--9698,
  \urlprefix\url{https://doi.org/10.1007/s11192-021-04147-y}

\bibitem[{Kruskal and Wallis(1952)}]{WilliamH.Kruskal.1952}
Kruskal WH, Wallis WA (1952) Use of ranks in one-criterion variance analysis.
  \emph{Journal of the American Statistical Association} 47(260):583--621,
  \urlprefix\url{https://doi.org/10.1080/01621459.1952.10483441}

\bibitem[{Kuhlman et~al.(2020)Kuhlman, Jackson, and
  Chunara}]{Kuhlman.27.02.2020}
Kuhlman C, Jackson L, Chunara R (2020) {No Computation without Representation:
  Avoiding Data and Algorithm Biases through Diversity}. In: Proceedings of the
  26th {ACM} {SIGKDD} Conference on Knowledge Discovery and Data Mining, {KDD}
  '20, p 3593, \urlprefix\url{https://doi.org/10.1145/3394486.3411074}

\bibitem[{Larivi{\`e}re et~al.(2018)Larivi{\`e}re, Macaluso, Mongeon, Siler,
  and Sugimoto}]{Lariviere.2018}
Larivi{\`e}re V, Macaluso B, Mongeon P, Siler K, Sugimoto CR (2018) Vanishing
  industries and the rising monopoly of universities in published research.
  \emph{PloS one} 13(8):e0202120,
  \urlprefix\url{https://doi.org/10.1371/journal.pone.0202120}

\bibitem[{Leydesdorff et~al.(2011)Leydesdorff, Bornmann, Mutz, and
  Opthof}]{Leydesdorff.2011}
Leydesdorff L, Bornmann L, Mutz R, Opthof T (2011) {Turning the tables on
  citation analysis one more time: Principles for comparing sets of documents}.
  \emph{Journal of the American Society for Information Science and Technology}
  62(7):1370--1381, \urlprefix\url{https://doi.org/10.1002/asi.21534}

\bibitem[{Littman(2021)}]{Littman.2021}
Littman LM (2021) Gathering strength, gathering storms: The one hundred year
  study on artificial intelligence: 2021 study panel report.
  \urlprefix\url{https://ai100.stanford.edu/2021-report/gathering-strength-gathering-storms-one-hundred-year-study-artificial-intelligence}

\bibitem[{Makridakis(2017)}]{Makridakis.2017}
Makridakis S (2017) {The forthcoming Artificial Intelligence (AI) revolution:
  Its impact on society and firms}. \emph{Futures} 90:46--60,
  \urlprefix\url{https://doi.org/10.1016/j.futures.2017.03.006}

\bibitem[{Mann and Whitney(1947)}]{MannWhitney1947}
Mann HB, Whitney DR (1947) {On a Test of Whether one of Two Random Variables is
  Stochastically Larger than the Other}. \emph{The Annals of Mathematical
  Statistics} 18(1):50--60, \urlprefix\url{http://www.jstor.org/stable/2236101}

\bibitem[{Robinson-Garcia et~al.(2014)Robinson-Garcia, Torres-Salinas, Zahedi,
  and Costas}]{RobinsonGarcia.2014}
Robinson-Garcia N, Torres-Salinas D, Zahedi Z, Costas R (2014) {New data, new
  possibilities: Exploring the insides of Altmetric.com}. \emph{El Profesional
  de la Informacion} 23:359--366,
  \urlprefix\url{https://doi.org/10.3145/epi.2014.jul.03}

\bibitem[{Salajegheh and Dayari(2019)}]{Salajegheh.2019}
Salajegheh M, Dayari S (2019) {Comparing the Citations Counts and Altmetrics of
  the Top Medical Science Journals in Scopus}. \emph{International Journal of
  Information Science and Management} 17:59--72,
  \urlprefix\url{https://ijism.ricest.ac.ir/article_698291_220563031da4cb3a012c9c0e4c86cd00.pdf}

\bibitem[{Sinha et~al.(2015)Sinha, Shen, Song, Ma, Eide, Hsu, and
  Wang}]{Sinha.2015}
Sinha A, Shen Z, Song Y, Ma H, Eide D, Hsu BJ, Wang K (2015) {An Overview of
  Microsoft Academic Service (MAS) and Applications}. In: Proceedings of the
  24th International Conference on World Wide Web, WWW'15, pp 243--246,
  \urlprefix\url{https://doi.org/10.1145/2740908.2742839}

\bibitem[{Smuha(2019)}]{smuha2019eu}
Smuha NA (2019) {The EU Approach to Ethics Guidelines for Trustworthy
  Artificial Intelligence}. \emph{Computer Law Review International}
  20(4):97--106, \urlprefix\url{https://doi.org/10.9785/cri-2019-200402}

\bibitem[{Spicer et~al.(2022)Spicer, Colcomb, and Kraft}]{spicer2022mind}
Spicer AJ, Colcomb PA, Kraft A (2022) Mind the gap: closing the growing chasm
  between academia and industry. \emph{Nature Biotechnology} 40(11):1693--1696,
  \urlprefix\url{https://doi.org/10.1038/s41587-022-01543-4}

\bibitem[{Tijssen(2004)}]{Tijssen.2004}
Tijssen RJ (2004) {Is the commercialisation of scientific research affecting
  the production of public knowledge? Global trends in the output of corporate
  research articles}. \emph{Research Policy} 33(5):709--733,
  \urlprefix\url{https://doi.org/10.1016/j.respol.2003.11.002}

\bibitem[{Vinuesa et~al.(2020)Vinuesa, Azizpour, Leite, Balaam, Dignum,
  Domisch, Fell{\"a}nder, Langhans, Tegmark, and Fuso~Nerini}]{vinuesa2020role}
Vinuesa R, Azizpour H, Leite I, Balaam M, Dignum V, Domisch S, Fell{\"a}nder A,
  Langhans SD, Tegmark M, Fuso~Nerini F (2020) The role of artificial
  intelligence in achieving the sustainable development goals. \emph{Nature
  communications} 11(1):1--10,
  \urlprefix\url{https://doi.org/10.1038/s41467-019-14108-y}

\bibitem[{Visser et~al.(2021)Visser, {van Eck}, and Waltman}]{Visser.2021}
Visser M, {van Eck} NJ, Waltman L (2021) {Large-scale comparison of
  bibliographic data sources: Scopus, Web of Science, Dimensions, Crossref, and
  Microsoft Academic}. \emph{Quantitative Science Studies} 2(1):20--41,
  \urlprefix\url{https://doi.org/10.1162/qss_a_00112}

\bibitem[{Wang et~al.(2020)Wang, Shen, Huang, Wu, Dong, and
  Kanakia}]{Wang.2020}
Wang K, Shen Z, Huang C, Wu CH, Dong Y, Kanakia A (2020) {Microsoft Academic
  Graph: When experts are not enough}. \emph{Quantitative Science Studies}
  1(1):396--413, \urlprefix\url{https://doi.org/10.1162/qss_a_00021}

\bibitem[{Wilkinson et~al.(2016)Wilkinson, Dumontier, Aalbersberg, Appleton,
  Axton, Baak, Blomberg, Boiten, da~Silva~Santos, Bourne
  et~al.}]{wilkinson2016fair}
Wilkinson MD, Dumontier M, Aalbersberg IJ, Appleton G, Axton M, Baak A,
  Blomberg N, Boiten JW, da~Silva~Santos LB, Bourne PE, et~al. (2016) {The FAIR
  Guiding Principles for scientific data management and stewardship}.
  \emph{Scientific data} 3(1):1--9,
  \urlprefix\url{https://doi.org/10.1038/sdata.2016.18}

\bibitem[{Williams(2022)}]{Williams.2022}
Williams K (2022) What counts: Making sense of metrics of research value.
  \emph{Science and Public Policy} 49:518–531,
  \urlprefix\url{https://doi.org/10.1093/scipol/scac004}

\bibitem[{Yan et~al.(2011)Yan, Tang, Liu, Shan, and Li}]{Yan.2011}
Yan R, Tang J, Liu X, Shan D, Li X (2011) {Citation Count Prediction: Learning
  to Estimate Future Citations for Literature}. In: Proceedings of the 20th ACM
  International Conference on Information and Knowledge Management, CIKM'11, p
  1247–1252, \urlprefix\url{https://doi.org/10.1145/2063576.2063757}

\bibitem[{Yang et~al.(2019)Yang, Rahardja, and Fr\"{a}nti}]{Outliers2019}
Yang J, Rahardja S, Fr\"{a}nti P (2019) Outlier detection: How to threshold
  outlier scores? In: Proceedings of the International Conference on Artificial
  Intelligence, Information Processing and Cloud Computing, AIIPCC'19,
  \urlprefix\url{https://doi.org/10.1145/3371425.3371427}

\bibitem[{Yu et~al.(2014)Yu, Yu, Li, and Wang}]{Yu.2014}
Yu T, Yu G, Li PY, Wang L (2014) Citation impact prediction for scientific
  papers using stepwise regression analysis. \emph{Scientometrics}
  101(2):1233--1252, \urlprefix\url{https://doi.org/10.1007/s11192-014-1279-6}

\bibitem[{Zhang et~al.(2021)Zhang, Mishra, Brynjolfsson, Etchemendy, Ganguli,
  Grosz, Lyons, Manyika, Niebles, Sellitto, Shoham, Clark, and
  Perrault}]{Zhang.09.03.2021}
Zhang D, Mishra S, Brynjolfsson E, Etchemendy J, Ganguli D, Grosz BJ, Lyons T,
  Manyika J, Niebles JC, Sellitto M, Shoham Y, Clark J, Perrault CR (2021) The
  {AI} index 2021 annual report. \emph{CoRR} abs/2103.06312,
  \urlprefix\url{https://arxiv.org/abs/2103.06312}

\bibitem[{Zhang et~al.(2019)Zhang, Wang, Zhao, {Ord{\'o}{\~n}ez de Pablos},
  Sun, and Xiong}]{Zhang.2019}
Zhang X, Wang X, Zhao H, {Ord{\'o}{\~n}ez de Pablos} P, Sun Y, Xiong H (2019)
  An effectiveness analysis of altmetrics indices for different levels of
  artificial intelligence publications. \emph{Scientometrics}
  119(3):1311--1344, \urlprefix\url{https://doi.org/10.1007/s11192-019-03088-x}

\end{thebibliography}

\end{document}